\documentclass[lettersize,journal]{IEEEtran}

\usepackage{amssymb}

\usepackage{algpseudocode}
\usepackage{algorithm}

\usepackage{graphicx}
\usepackage{subfigure}
\usepackage{array}
\usepackage{amssymb}
\usepackage{subfig}
\usepackage{tabulary}
\usepackage{booktabs}
\usepackage{bm}
\usepackage{amsmath}
\usepackage{numprint}
\npthousandsep{,}
\usepackage{multirow}
\usepackage{threeparttable}
\usepackage{lscape}
\usepackage{rotating}
\usepackage{color}

\usepackage{float}
\usepackage{caption}

\usepackage{adjustbox}
\usepackage[utf8]{inputenc}

\usepackage{amsmath}
\usepackage{makecell} 
\usepackage{booktabs} 

\begin{document}

\title{\begin{LARGE}Interpretable Recognition of Fused Magnesium Furnace Working Conditions with Deep Convolutional Stochastic Configuration Networks  \end{LARGE}}

\author{ Weitao Li, Xinru Zhang, Dianhui Wang, Qianqian Tong, and Tianyou Chai

\thanks{This work was supported by National Key R\&D Program of China under Grant (2018AAA0100304), National Natural Science Foundation of China (62173120, 62103092), Anhui Provincial Natural Science Foundation (2108085UD11) and 111 Project (BP0719039). (\emph{Corresponding author: Dianhui Wang.})}

\thanks{Weitao Li, Xinru Zhang, and Qianqian Tong are with the Department of Electric Engineering and Automation of Hefei University of Technology, Hefei 230009, China  (e-mail: wtli@hfut.edu.cn; 2021110400@mail.hfut.edu.cn; xrzhang@mail.hfut.edu.cn).}

\thanks{Dianhui Wang is with the Institute of Artificial Intelligence, China University of Mining and Technology, Xuzhou 221116, China  (e-mail: dh.wang@deepscn.com).}

\thanks{Tianyou Chai is with the State Key Laboratory of Synthetical Automation for Process Industries, Northeastern University, Shenyang 110004, Liaoning, China (e-mail: tychai@mail.neu.edu.cn).}
}

\markboth{Journal of \LaTeX\ Class Files,~Vol.~14, No.~8, May~2024}%
{Shell \MakeLowercase{\textit{et al.}}: A Sample Article Using IEEEtran.cls for IEEE Journals}


\maketitle

\begin{abstract}
To address the issues of a weak generalization capability and interpretability in working condition recognition model of a fused magnesium furnace, this paper proposes an interpretable working condition recognition method based on deep convolutional stochastic configuration networks (DCSCNs). Firstly, a supervised learning mechanism is employed to generate physically meaningful Gaussian differential convolution kernels. An incremental method is utilized to construct a DCSCNs model, ensuring the convergence of recognition errors in a hierarchical manner and avoiding the iterative optimization process of convolutional kernel parameters using the widely used backpropagation algorithm. The independent coefficient of channel feature maps is defined to obtain the visualization results of feature class activation maps for the fused magnesium furnace. A joint reward function is constructed based on the recognition accuracy, the interpretable trustworthiness evaluation metrics, and the model parameter quantity. Reinforcement learning (RL) is applied to adaptively prune the convolutional kernels of the DCSCNs model, aiming to build a compact, highly performed and interpretable network. The experimental results demonstrate that the proposed method outperforms the other deep learning approaches in terms of recognition accuracy and interpretability.
\end{abstract}

\begin{IEEEkeywords}
Deep convolutional stochastic configuration networks, Gaussian differential convolution kernel, Class activation mapping map, Interpretability, Fused magnesium furnace working condition recognition.
\end{IEEEkeywords}

\section{Introduction}
\IEEEPARstart{M}{agnesium} oxide, as the main component of fused magnesium sand (also known as fused magnesia), is an alkaline refractory raw material widely used in aerospace, nuclear furnaces, electronics, and other fields. As the world's largest producer and supplier of fused magnesium, China faces challenges such as low grade, large composition fluctuations, and complex mineral composition in its magnesite ore. Therefore, a unique three-phase alternating current fused magnesium furnace is required for smelting. The smelting process of an fused magnesium furnace involves simultaneous feeding and smelting. Raw materials are poured into the furnace by a machine, and high-temperature arcs in the furnace heat the raw materials to generate magnesium oxide crystals \cite{9760087, 9380574, 8563103}. 

To ensure the quality of electric fused magnesia, it is necessary to monitor the production process and prevent possible abnormal working conditions during smelting, including underburn, overheating, and abnormal exhaust \cite{6880336}. The underburn condition is caused by impurities and complex minerals in the raw materials, resulting in incomplete combustion in the fused magnesium furnace. During this condition, the furnace wall in certain areas becomes red and bright. The overheating condition is characterized by a brighter flame at the furnace mouth, which may produce adverse substances such as magnesium smoke and magnesium oxide. The abnormal exhaust working condition involves the ejection of high-temperature molten metal from the furnace mouth, resulting in drastic changes in current. When these abnormal working conditions occur, they need to be promptly detected and addressed to avoid energy waste, reduction of the utilization rate and grade of magnesia sand, furnace lining burn-through, and potential hazards to operators due to raw material leakage. Presently, the inability to directly observe the internal molten pool in the furnace due to the presence of burning flames at the furnace mouth renders the diagnosis of abnormal working conditions in magnesium furnaces contingent upon manual decision-making \cite{9917360}.  However, each inspector is responsible for multiple furnaces, and subjective factors such as their experience, sense of responsibility, and labor intensity, as well as objective factors such as high temperature, noise, dust, water mist, and inherent white spots on the furnace wall in a complex firing environment, make it difficult to make manual decisions. This can easily lead to missed or erroneous inspections, resulting in irreversible losses such as the burn-through of fused magnesium furnaces and difficulties in meeting the real-time inspection requirements for operation and maintenance.

In recent years, with the continuous development of artificial intelligence technology, the use of machine learning and deep learning for the abnormal working condition diagnosis of fused magnesium furnaces has received widespread attention. These methods collect and process production data from the furnaces, extract various feature parameters, and utilize algorithms for the automatic diagnosis of abnormal working conditions. For example, the work in \cite{LI2018178} proposes a magnesium furnace abnormal working condition diagnosis method based on Bayesian networks, which introduces transfer learning to address the problem of limited abnormal working condition data. The work in \cite{9610130} applies semi-supervised learning to automatically label unlabeled current data and train classifiers constructed within a semi-supervised learning framework. The work in \cite{8888888} combines convolutional neural networks (CNNs) and long short-term memory (LSTM) networks to extract the spatial and temporal features of fused magnesium furnaces for underburn condition recognition. The work in \cite{BU2022357} uses convolutional networks, bi-directional long short-term memory networks (Bi-LSTM), and stacked autoencoders to respectively extract image, sound, and current features of fused magnesium furnaces, and then combines these different features to train a condition classifier. However, the unique operational conditions in fused magnesium furnaces present a challenge for the generalization capabilities of recognition models. Models that have been otherwise well-trained exhibit a marked decline in performance when exposed to these datasets, revealing a significant weakness in their ability to generalize.

Although deep learning has been widely used in the field of image analysis, there are still some challenges that need to be addressed. The black box nature of deep neural networks is essentially due to the inconsistency between the features learned by internal neurons and the semantic concepts understood by humans. The interpretability of deep learning refers to the extent to which the decision-making process of a model can be clearly understood and explained. It is an important indicator of the level of understanding of the model and the degree of trust in its decisions. Currently, the interpretability of deep learning models can be presented in different ways \cite{10412652,7552539}, including feature visualization, analysis visualization, local and global interpretability, and the design of interpretable network structures. Of these, interpretable network structures can help users understand the model. They not only allow observation of the model's prediction results but also provide insights into the reasons behind the model's decisions. In the case of model errors, users can make adjustments to the model themselves \cite{10337787}. Traditional methods for network construction usually rely on iterative trial and error based on human prior knowledge and experience, which consumes a considerable amount of time. In recent years, researchers have proposed some automatic network structure search methods \cite{10436701}. The work in \cite{9982412} introduces a graph-based neural architecture encoding scheme (GATES), which employs an intuitive modeling approach that mirrors the data processing flow of neural networks. This enhances the representation capability of neural architectures, leading to a substantial improvement in predictor performance across various cell-based search spaces. The work in \cite{8966988} proposes a block-wise network generation pipeline (BlockQNN), which employs the Q-Learning paradigm with epsilon-greedy exploration strategy to automatically construct the network structure. However, these methods have a large search space and low computational efficiency. Additionally, training neural networks using backpropagation and gradient descent suffers from issues such as weight initialization, local minima, and the sensitivity of learning performance to learning rates, which also limit the performance of deep learning models. Therefore, it is crucial to research a fast and efficient automatic construction method for a network structure that possesses interpretability.

To address issues such as the long training time, susceptibility to local minima, and the difficulty in determining the number of hidden layer nodes closely related to the learning capability in neural networks, random learning algorithms have emerged. This algorithm randomly assigns input weights and biases and calculates output weights using least squares estimation. Distinguished from the random vector functional link networks \cite{Pao199276}, stochastic configuration networks (SCNs) are featured by using a supervisory mechanism to assign the random parameters and adaptively select the scope of the random parameters, ensuring the universal approximation property of the constructed randomized learners at algorithmic level. The work in \cite{8767029} introduces a two-dimensional stochastic configuration network (2DSCNs) that can directly process two-dimensional data, improving the generalization performance of SCNs in image data modeling. The work in \cite{8489695} proposes a deep stochastic configuration network (DeepSCNs) where hidden nodes in each layer are connected to the output, allowing the network to learn richer feature representations. Due to the merits of the SCNs framework, it has received considerable attention in the field of industrial data analysis \cite{ 10226533, 10629070, 10251649, 8998381, 9345932}. 

Although SCNs construct randomized learners with the universal approximation property, they have a limited ability to extract image features, making them inadequate for representation of interest image information. In the construction process of CNNs, the network width is usually preset, i.e., the number of convolutional kernels in each layer. Due to the possible correlation between the convolutional kernels, coupled with the backpropagation to iteratively update the parameters of the convolutional kernels, the computation load for the training of CNNs is massive \cite{lecun2015deep}. The work in \cite{tong} proposes an interpretable method for recognizing working conditions in fused magnesium furnace based on deep convolutional stochastic configuration networks (DCSCNs). This approach utilizes a supervised learning mechanism to generate Gaussian differential convolution kernel with physical meaning, and uses an incremental method to construct a deep convolutional neural network to ensure that the recognition error converges step by step and avoid the process of backpropagation algorithm to iteratively find the optimal convolutional kernel parameters. Additionally, channel feature map independence coefficients and interpretable credibility indexes are defined to visualize the activation mapping maps for fused magnesium furnace feature classes. Due to possible redundant information in the associated convolutional kernels, the pruning optimization of the convolutional kernels in the network is necessary. The work in \cite{9847369} proposes a pruning pattern that removes consecutive $N$ output kernels sharing the same input channel index, preserving model accuracy while achieving substantial speedups on general CPUs. The work in \cite{9384353} proposes a channel pruning method for the compression of deep neural networks, which introduces additional loss functions then performs layer-wise channel selection to recognize channels and kernels that meaningfully contribute to the discriminative capability. Applying reinforcement learning (RL) to prune the convolutional kernels enables the end-to-end learning of the optimal compressed feature combinations, thereby reducing network size and computational costs \cite{CHUNG2024121265}.

In summary, this study addresses the shortcomings related to the generalization and interpretability of recognition models in the specific operational environments of fused magnesium furnaces. Leveraging the explainable architecture of SCNs, an interpretable recognition method of fused magnesium furnace working conditions with deep convolutional stochastic configuration networks ((DCSCNs) is proposed. The major contributions of this study are as follows:

1) To overcome the limitations of traditional incremental convolutional networks, a stochastic configuration approach is used for the first time to construct a deep CNNs. The architecture starts with a single-layer network with a single convolutional kernel and incrementally expands until it meets the stopping criteria for iterations. Physically interpretable Gaussian differential convolutional kernel parameters are automatically configured through a data-correlated parameter selection strategy. This approach eliminates the need for iterative optimization of convolutional kernel parameters based on backpropagation algorithm, and ensures that the recognition error converges step by step. 
A proof of global convergence capabilities for the deep CNNs is also provided, further substantiating the model interpretability.

2) The feature maps generated by DCSCNs are upsampled to the size of the input image using bilinear interpolation. After overlaying the upsampled feature maps with the original image, they are re-inputted to the DCSCNs under the current level of stochastic convolutional configuration layers to obtain scores indicating the membership of the multi-modal conditions to different categories. The independent coefficient of channel feature maps is defined to measure the independence score of different channel features. The category scores, channel feature independence scores, and corresponding channel feature maps are linearly combined to generate class activation maps. By overlaying these maps onto the original image, the feature visualization results of the fused magnesium furnace under the current level condition can be obtained. An evaluative index for interpretable trustworthiness is defined to measure the consistency between the interpretability results and the ground truth.

3) The generated Gaussian differential convolutional kernels are sorted based on the independent scores of channel features. A joint reward function is defined, which includes an interpretable trustworthiness evaluative index, recognition accuracy, and model parameter quantity. RL is employed to adaptively prune the Gaussian differential convolutional kernel set to adjust the network width, obtaining optimal interpretable condition recognition results.

The remainder of this paper is organized as follows: Section II introduces the interpretable fused magnesium furnace working condition recognition model, outlining its three core components: the training, feedback, and testing layers, and explaining their synergistic interactions to improve performance. Section III describes an interpretable working condition recognition method based on DCSCNs, incorporating data augmentation, and a novel convolutional kernel parameter selection strategy. Section IV provides a rigorous theoretical proof of the convergence of the proposed DCSCNs. Section V leverages an activation mapping method based on feature independence to assess the interpretability of working condition recognition results in the fused magnesium furnace. Section VI proposes an adaptive pruning mechanism for convolutional kernels based on RL, which optimizes network width by ranking kernels based on feature independence scores. Section VII details the data, evaluation metrics, and experimental results, covering the DCSCNs performance, the interpretability evaluation, the convolutional kernel adaptive pruning, ablation studies, and performance comparisons. The final section summarizes the contributions, strengths, and potential directions for future research.

\section{Interpretable fused magnesium furnace working condition recognition model}
The proposed interpretable working condition recognition model for a fused magnesium furnace based on DCSCNs utilizes a three-layer structure for information coupling, including the training layer, feedback layer, and testing layer. The model structure is illustrated in Figure 1.

\begin{figure*}[!t]
\centering
\includegraphics[width=6in]{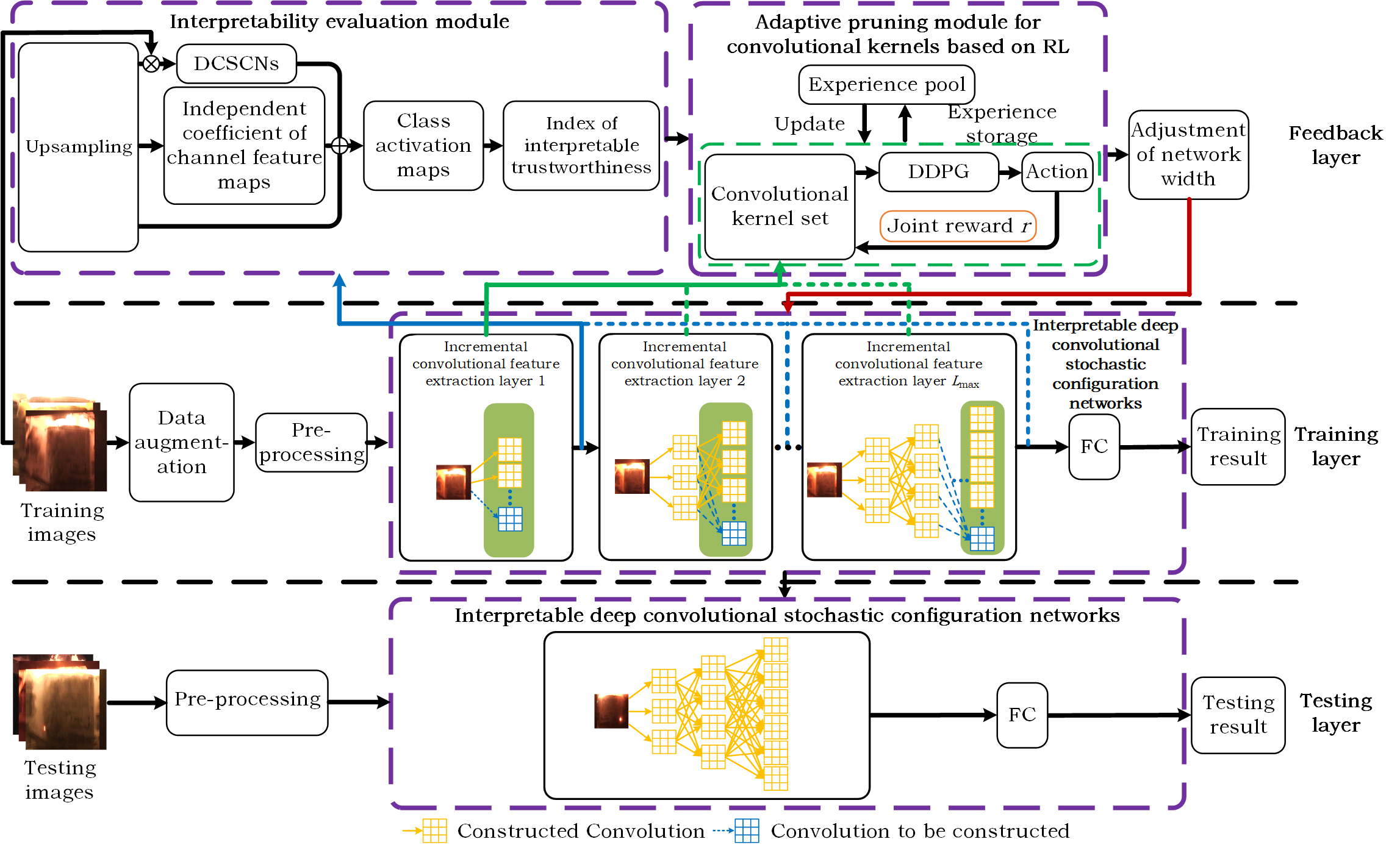}
\caption{Structure of interpretable fused magnesium furnace working condition recognition model based on deep convolutional stochastic configuration networks.}
\label{fig_1}
\end{figure*}

The training layer consists of data augmentation, preprocessing, and DCSCNs modules. Initially, fused magnesium furnace training images undergo data augmentation and preprocessing to enlarge the dataset. These processed samples are then fed into the DCSCNs. A supervised mechanism is opted for the adaptive selection of new convolutional kernel parameters to avoids the need for the iterative backpropagation for updating convolutional kernel parameters. This process starts with a single-kernel and single-layer configuration (represented as "Incremental convolutional feature extraction layer 1", the yellow box as the constructed convolution, and the blue box as the convolution to be constructed). The training set is passed through this layer to obtain the feature map. The feature map is then used for classifying the working conditions of the fused magnesium furnace through a fully connected layer. If the recognition error meets the preset criteria, training is halted. Otherwise, the number of single-layer convolutional kernels continues to increase one by one at a time (indicated as the second yellow box of the incremental convolutional feature extraction layer 1) until it reaches the maximum allowable kernels ($C_1$) for that layer. If the error still does not meet the criteria, a second incremental convolutional feature extraction layer is added (represented as "Incremental convolutional feature extraction layer 2"). This process is repeated until the error converges or the maximum number $L_{\mathrm{max}}$ of convolutional layers is reached.

The feedback layer encompasses an interpretability assessment module and a network width adaptive adjustment module based on RL. After the construction of $L_{\mathrm{max}}-1$ layers in the DCSCNs, the extracted feature maps are fed into the feedback layer (indicated by the blue arrow). These are upsampled to the original input image size and are superimposed with the original image. The superimposed image is re-inputted into the incremental convolutional network to obtain class scores. Channel feature independence scores are calculated, and a linear combination of the class scores, channel feature independence scores, and feature maps is used to construct a class activation mapping (CAM) and a measure index of interpretability trustworthiness. Meanwhile, the Gaussian differential convolutional kernels generated are then fed into a RL pruning module (indicated by the green arrow). The kernels are ranked based on their channel feature independence scores, and a joint reward function that combines accuracy, a measure index of interpretability trustworthiness, and the volume of network parameters is formulated. Convolutional kernels for each layer of the network are adaptively pruned through RL, enabling the self-adaptive adjustment of network width (indicated by the red arrow).

In the testing layer, the constructed interpretable deep stochastic configuration convolutional network is used to obtain the optimal recognition results for the test samples of fused magnesium furnace working conditions.

\section{Interpretable working condition recognition method based on deep convolutional stochastic configuration networks}
\subsection{Data augmentation and data preprocessing }
In the production process of fused magnesium furnaces, there are relatively few instances of abnormal working conditions. This leads to an imbalance issue in the image samples, which can cause over-fitting of the learner. To address this problem, this paper employs a non-generative approach to augment the image data. Techniques such as horizontal flipping, contrast and brightness adjustment, and adding noise are applied to increase the diversity of the data and alleviate over-fitting, enabling the learner to better adapt to the variations in magnesium furnace images across different scenarios.

Horizontal flipping can be described as follows:

\begin{equation}
\label{deqn_ex1a}
I(x,y)=I'(w-x-1,y)
\end{equation}
where, $I'(x,y)$ represents the pixel value of the original image at coordinates $(x,y)$, $I(x,y)$ denotes the pixel value of the data enhanced image at $(x,y)$, and $w$ denotes the width of the original image.

Contrast and brightness adjustment can be described as follows:

\begin{equation}
\label{deqn_ex1a}
I(x,y)=1.5×(I'(x,y)-0.5)+0.5
\end{equation}

To prevent pixel value overflow, each value in the original image is subtracted by 0.5, multiplied by the contrast enhancement coefficient of 1.5, and then added by 0.5. This simultaneously enhances the brightness and contrast of the image.

Adding Gaussian noise can be described as follows:

\begin{equation}
\label{deqn_ex1a}
I(x,y)=I'(x,y)+N(0,\eta^2 )
\end{equation}
where, $N(0,\eta^2)$ represents Gaussian noise with a mean of 0 and a standard deviation of $\eta ^2$. By adjusting the value of $\eta $, the intensity of the noise can be controlled.

The collected images may contain some information unrelated to the fused magnesium furnace. To reduce the impact of such information, the center of the image is cropped to $1080\times 1080$ and resized to $256\times 256$. The input values are normalized to the range of $[-1,1]$ for subsequent image processing convenience.

\subsection{Deep convolutional stochastic configuration networks}
To address the issues of structure design, hyperparameter tuning, and interpretability in CNNs, this paper proposes an efficient method for constructing interpretable CNNs as shown in Figure 2, called the DCSCNs, inspired by the stochastic configuration network \cite{8013920,LI202261,tong}. This strategy starts from a single convolutional kernel in a single layer and incrementally generates new stochastic convolutional kernels with strong interpretability to construct the CNNs. It overcomes the tediousness of manually designing network structures and tuning hyperparameters using traditional gradient descent methods. The supervised learning mechanism is employed in the selection of convolutional kernel parameters, ensuring the global approximation capability of the deep learning model.

\begin{figure*}[!t]
\centering
\includegraphics[width=6in]{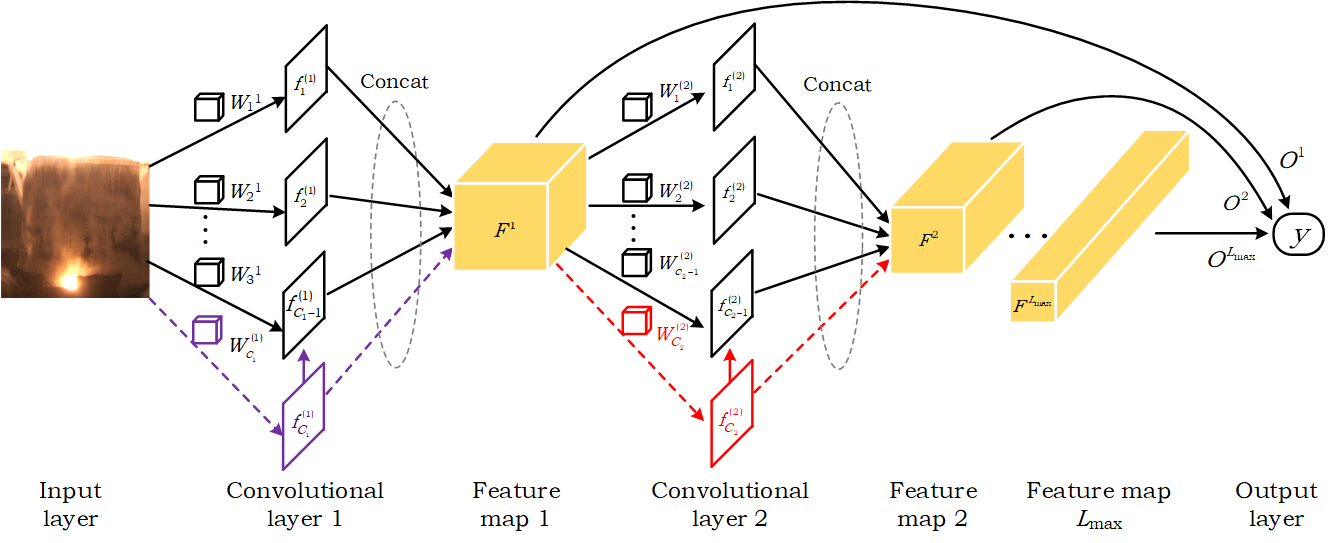}
\caption{Deep convolutional stochastic configuration network structure diagram.}
\label{fig_2}
\end{figure*}

\subsubsection{Generation strategies of randomized configuration convolutional kernel}

The stochastic configuration convolutional generation strategy adaptively determines the range of convolutional kernel parameters and randomly generates new kernel parameters within this range, including weights and biases. The deep neural network is constructed starting from zero convolutional kernels, and interpretable new convolutional kernels are generated using the supervised learning mechanism, ensuring the improvement of network performance. Finally, the output weights of the network are updated using the least squares method, and the generation of convolutional kernels stops when the network error is below the predefined threshold.

Let $F=[f_1,f_2,…,f_m]:\mathbb{R}^{d_1\times d_2\times d_3} \to \mathbb{R}^m$ be a set of real-valued functions, where the $L_2$ norm is defined as:

\begin{equation}
\label{deqn_ex1a}
\small
\left \| F \right \| =\left (  {\textstyle \sum_{q=1}^{m}}  \int_{D}\left | f_q(x)^2 \right | dx \right )^{1/2}< \infty 
\small
\end{equation}

The inner product between the real-valued functions $F$ and $G=[g_1,g_2,…,g_m]:\mathbb{R}^{d_1\times d_2\times d_3} \to \mathbb{R}^m$ can be represented as:

\begin{equation}
\label{deqn_ex1a}
 \langle F, G\rangle=\sum_{q=1}^m\left\langle f_q, g_q\right\rangle=  \sum_{q=1}^m \int_D f_q(x) g_q(x) d x
\end{equation}

Given an input matrix $I$ and a convolutional kernel $K$ of size $\rho \times n$, the element at position $(i, j)$ in the output matrix $S$ after the cross-correlation operation is defined as:

\begin{equation}
\label{deqn_ex1a}
 S(i, j)=(I * K)(i, j)=\sum_\rho \sum_n I(i+\rho, j+   n) K(\rho, n) \quad 
\end{equation}
where, $*$ denotes the cross-correlation operator and $I(i+\rho,j+n)K(\rho,n)$  represents the element at position $(i+\rho,j+n)$ in the input matrix $I$ multiplied by the element at position $(\rho,n)$ in the convolutional kernel $K$.

Given a target function $F: \mathbb{R}^d \to \mathbb{R}^m$, assuming a DCSCNs consists of $L_{\mathrm{max}}$ convolutional layers, each layer with $C_1, \ldots, C_l, \ldots, C_{L_{\max }}\left(l \in\left[1, L_{\max}\right]\right)$ convolutional kernels with a size of $k\times k$, the DCSCNs can be represented as:

\begin{equation}
\label{deqn_ex1a}
 F^l(x)=\sum_{l=1}^{L_{\max}} \sum_{C=1}^{C_l} O_C^l A_C^l\left(W_C^l, b_C^l, I^l\right) 
\end{equation}
where, $W_C^l$ and $b_C^l$ are the parameters of the $C^{th}$ convolutional kernel in the $l^{th}$ layer, $W_C^l=$ $\left[\begin{array}{ccc}W_{C,[1,1]}^l & \cdots & W_{C,[1, k]}^l \\ \vdots & \ddots & \vdots \\ W_{C,[k, 1]}^l & \cdots & W_{C,[k, k]}^l\end{array}\right]$, $I^l$ is the input of the $l^{th}$ layer, $A_C^l$ represents the convolutional function, including convolutional operation, activation function, and downsampling operation, $O_C^l$ is the output weight, and the error $e^l=F-F^l=\left[e_1^l, e_2^l, \ldots, e_m^l\right]$.

Assuming the function space $\operatorname{span}(\Gamma)$ composed of $\Gamma $ is dense in the $L_2$ space, $\forall A \in \Gamma$, $0<\|A\|<b$, where $b \in \mathbb{R}^{+}$. Given $p>0$ and a non-negative contraction sequence $\left\{u_C\right\}$, $\left\{u_C\right\}$ satisfying $\lim _{C \rightarrow+\infty} u_C=$ 0 and $\sum_{C=1}^{\infty} u_C=\infty$, the parameters of the $C^{th}$ convolutional kernel in the $l^{th}$ layer are randomly generated if they satisfy:

\begin{equation}
\quad\left\langle e_{C, q}^l, A_C^l\right\rangle^2 \geq p u_C b^2\left\|e_{C-1, q}^l\right\|^2
\end{equation}
where, $q=1,2, \ldots, m$. Then, the condition $\lim _{l \rightarrow+\infty}\left\|F-F^l\right\|=0$ holds; otherwise, the parameters of the $C^{th}$ convolutional kernel in the $l^{th}$ layer are regenerated. When the number of convolutional kernels in the $l^{th}$ layer increases to $C_l$  if the error $e^l$ is still greater than the predetermined value, a new convolutional kernel is added as the first kernel of the $l+1^{th}$ layer, and it satisfies:

\begin{equation}
\label{deqn_ex1a}
\left\langle e_{C_l, q}^l, A_1^{l+1}\right\rangle^2 \geq \xi u_{C_l} b^2\left\|e_{C_l-1, q}^l\right\|^2
\end{equation}

Then, the condition $\lim _{l \rightarrow+\infty}\left\|F-F^l\right\|= 0$; otherwise, the parameters of the first convolutional kernel in the $(l+1)^{th}$ layer are regenerated until the termination condition for kernel augmentation is met.

Therefore, the construction problem of a DCSCNs can be described as follows: Given training image data $X=\left\{x_1, x_2, \ldots, x_N\right\}$, where $x_i \in \mathbb{R}^{d_1 \times d_2 \times d_3}$, and the corresponding outputs $Y=\left\{y_1, y_2, \ldots, y_N\right\}$, where $y_i \in \mathbb{R}^m$ represents the image category label. Let $I_t^l$ denote the $t^{th}$ channel of the $l^{th}$ layer convolution of input data $I^l$, for $t=1,2, \ldots, C_{l-1}$. The feature map $M_C^l$, representing the output of the $C^{th}$ convolutional kernel in the $l^{th}$ layer, can be expressed as:

\begin{equation}
M_C^l=g\left(\sum_{t=1}^{C_{l-1}} W_C^l * I_t^l+b_C^l\right) 
\end{equation}
where, $g(\cdot)$ is the activation function, and $M_C^l$ has dimensions $H \times W$. After max pooling, $M_C^l$ obtains the downsampled feature map $A_C^I$:

\begin{equation}
 A_C^l=\max _{m=1}^{k-1} \max _{n=1}^{k-1} M_{C, \Psi, \Omega}^l 
\end{equation}
where, $\quad \Psi \in[h, h+m], h=1,2, \ldots$, $H$ represents the length range of the feature map, and $\Omega \in[w, w+n]$ with $\quad w=1,2, \ldots, W$ represents the width range of the feature map.

Based on Equation (11), the feature map set $A^l=\left[A_1^l, A_2^l, \ldots, A_C^l\right]$ of the $l^{th}$ layer convolution can be obtained. Multiplying $A_C^l$ by the output weights $O_C^l$ of the convolutional network yields the output $F^l$ for the $C^{th}$ convolutional kernel in the $l^{th}$ layer:
\begin{equation}
 F^l=\sum_l \sum_{C=1}^{C_l} O_C^l A_C^l 
\end{equation}

The output weights $O_C^l$ are updated using the least squares method:
\begin{equation}
\begin{gathered}
O_C^l=\left[O_{C, 1}^l, \ldots, O_{C, m}^l\right]= 
\arg \min _O\left\|Y-\sum_l \sum_{C=1}^{C_l} O_C^l A_C^l\right\|
\end{gathered}
\end{equation}

The error $e_C^l$ for the $C^{th}$ convolutional kernel in the $l^{th}$ convolutional layer is defined as:
\begin{equation}
e_C^l=Y-F^l=\left[e_{C, 1}^l, e_{C, 2}^l, \ldots, e_{C, m}^l\right] 
\end{equation}

If the $L_2$ norm $\left\|e_C^l\right\|_2$ of $e_C^l$ is greater than the desired error limit $\bar{e}$, new parameters for the convolutional kernel with weights $W_{C+1}^l$ and biases $b_{C+1}^l$ are generated until the stopping criterion is met.

\subsubsection{Convolutional kernel parameter selection strategy}

The construction of convolutional kernels in deep learning models directly affects the correlation between the learned model and input data. In a deep convolutional network with good performance, the convolutional kernels should follow a Gaussian distribution \cite{BAILON2006383}. The highlight regions of the furnace wall and furnace mouth exhibit distinct texture characteristics, with significant brightness variations at the boundaries. Gaussian differential convolutional kernels, which involve Gaussian smoothing and differencing operations on images, can enhance texture information to distinguish image regions where texture is weakened or lost due to fog occlusion. Therefore, Gaussian differential convolutional kernels are selected for the DCSCNs, defined as follows \cite{tong}:

\begin{equation}
\psi(x, y)=\frac{1}{2 \pi}\left[e^{-\frac{x^2+y^2}{2 \xi^2}}-\frac{1}{r} e^{-\frac{x^2+y^2}{2 r^2 \xi^2}}\right]
\end{equation}
where, $\xi$ represents the standard deviation, controlling the width range of the convolutional kernel, and $r$ represents the scale factor. Smaller $r$ values can detect fine edges and details, while larger $r$ values can detect rough edges and contour structures. Different combinations of $\xi$ and $r$ yield different information about the magnesium furnace conditions.

Unlike traditional methods that iteratively update convolutional kernel parameters using gradient descent algorithms, this study adopts a strategy of randomly selecting parameters to generate different convolutional kernels, where $\xi$ and $r$ follow uniform distributions. The weights $W_C^l$ of the convolutional kernels follow a Gaussian differential distribution as shown in Equation (15), and the biases $b_C^l$ are  selected from a uniform distribution as shown in the following equation:
\begin{equation}
b_C^l \sim U(0,1)
\end{equation}

By generating $T_{\max }$ candidate Gaussian differential convolutional kernels based on Equation (15) and Equation (16), the convergence score $\sigma_{C,q }$ of each candidate convolutional kernel can be evaluated using Equation(17):
\begin{equation}
\sigma_{C, q}^l=\frac{\left(\left(e_{C, q}^l\right)^{\mathrm{T}} A_{C, q}^l\right)^2}{\left(A_{C, q}^l\right)^{\mathrm{T}} A_{C, q}^l}-(\xi+r) u_C\left(e_{C, q}^l\right)^{\mathrm{T}} e_{C, q}^l
\end{equation}
where, $A_{C, q}^l=\left[A_{C, 1}^l, A_{C, 2}^l, \cdots, A_{C, m}^l,\right]$. Candidate Gaussian differential convolutional kernels satisfying Equation (18) are retained:
\begin{equation}
 \quad \sum_{q=1}^m \sigma_{C, q}^l>0 \quad 
\end{equation}

The candidate Gaussian differential convolutional kernel with the highest convergence score $\sum_{q=1}^m \sigma_{C, q}^l$ is selected as the $C^{th}$ convolutional kernel in the $l^{th}$ layer. If none of the $T_{\max}$ candidate Gaussian differential convolutional kernels meet the requirements, new $\xi$ and $r$ values are selected based on Equation (15) to generate a new Gaussian differential convolutional kernel. This process continues until a convolutional kernel in the $l^{th}$ layer satisfies the condition in Equation (18).

\subsubsection{Proof of convergence of deep convolutional stochastic configuration networks}

Assuming that the number of convolutional kernels in the $l^{th}$ layer of DCSCNs is $C_l$, it can be derived that the $l^{th}$ layer convolution satisfies \cite{tong}:

\begin{equation}
\left\|e_{C_l}^l\right\|^2 \leq(\xi+r) u_{C_l}\left\|e_{C_{l-1}}^l\right\|^2 
\end{equation}

The output error $e_1^{l+1}$ of the first Gaussian differential convolutional kernel generated in the $l+1^{th}$ layer convolution is given by:
\begin{equation}
\quad\left\|e_1^{l+1}\right\|^2=\left\|e_{C_l}^l-O_1^{l+1} A_1^{l+1}\right\|^2
\end{equation}
where, $O_1^{l+1}=\left[O_{1,1}^l, O_{1,2}^l, \ldots, O_{1, m}^l\right]$. Let $\bar{O}_1^{l+1}=$ $\left\lceil\bar{O}_{1,1}^l, \bar{O}_{1,2}^l, \ldots, \bar{O}_{1, m}^l\right\rceil$ represent the optimal output weights. Then $\bar{O}_{1, q}^{l+1}$ can be expressed as:
\begin{equation}
\bar{O}_{1, q}^{l+1}=\frac{\left\langle e_{C_l, q}^l, A_1^{l+1}\right\rangle}{\left\|A_1^{l+1}\right\|^2}
\end{equation}

By transforming $\left\|e_1^{l+1}\right\|^2$ using Equations (18-21), we have:
\begin{equation}
\begin{array}{r}
\left\|e_1^{l+1}\right\|^2 \leq\left\|e_{C_l}^l-\bar{O}_1^{l+1} A_1^{l+1}\right\|^2 \\ 
=\left\|e_{C_l}^l\right\|^2-\sum_{q=1}^m \bar{O}_{1, q}^{l+1} \leq\left\|e_{C_l}^l\right\|^2 \\
\end{array}
\end{equation}

Thus 

\begin{equation}
\begin{array}{r}
\left\|e_{C_{l+1}}^{l+1}\right\|^2 \leq\left\|e_1^{l+1}\right\|^2 \leq\left\|e_{C_l}^l\right\|^2 \\ \leq(\xi+r) u_1^{\left(\sum_l C_l\right)-1} \left\|e_1^1\right\|^2 
\end{array}
\end{equation}

From this, it can be proven that the output error $e_1^{l+1}$ of the DCSCNs monotonically decreases, indicating the convergence of the network.

\subsection{Interpretability evaluation}

To ensure the reliability and interpretability of the recognition results for working conditions in the fused magnesium furnace, it is important for operators to understand the trustworthy decision-making process of deep learning models. The DCSCNs ensures global convergence to obtain reliable recognition results \cite{tong}. The Gaussian distribution convolutional kernels generated based on supervised learning have a strong correlation with the input data, ensuring the interpretability of the model mechanism. Class activation mapping visualization, as a visual interpretability verification, represents the distribution of importance in the data. It utilizes the gradient information of the last convolutional layer as the category score for channel feature maps and calculates the category importance by weighting the feature maps. Finally, the feature activation mapping is superimposed on the original image to obtain the interpretable visualization of feature data as shown in Figure 3.

\begin{figure}[!t]
\centering
\includegraphics[width=3in]{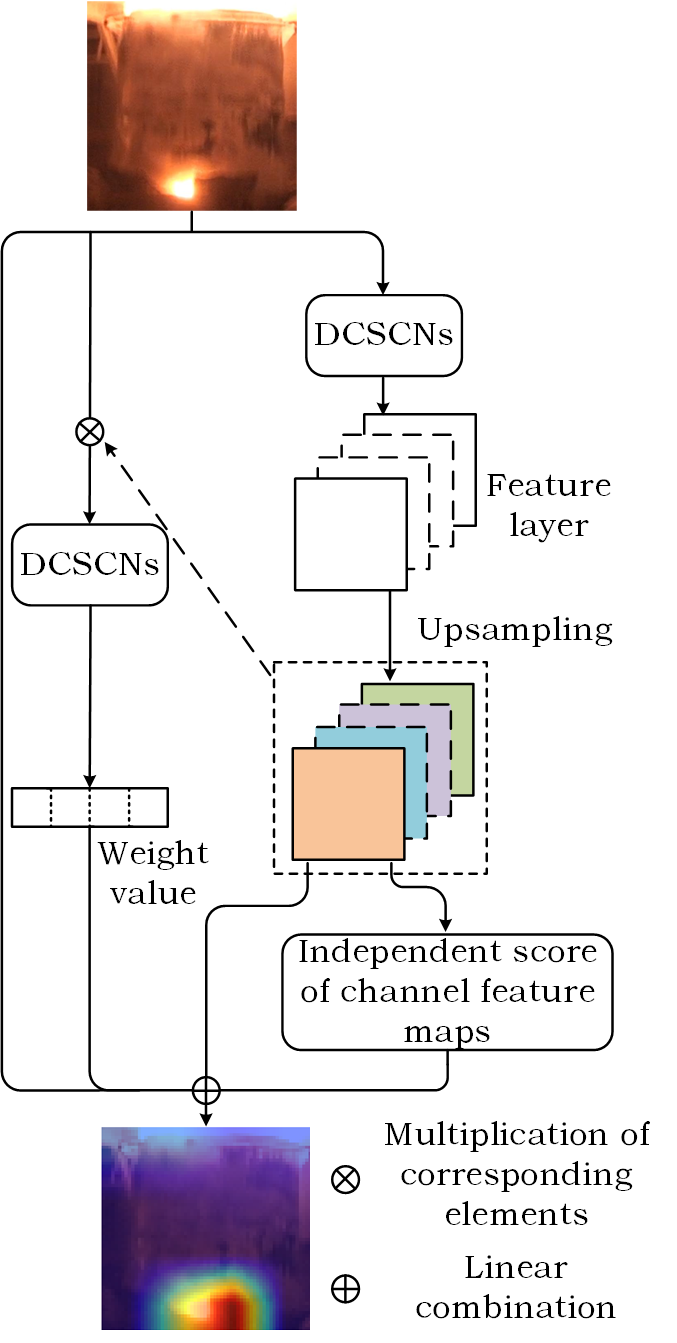}
\caption{Schematic diagram of the class activation mapping based on feature map independence scores.}
\label{fig_3}
\end{figure}

The feature maps set $A^l$ in the $l^{th}$ layer is upsampled to the original input size using bilinear interpolation and normalized to the range $[0,1]$. The $\varrho^{th}$ channel feature map $A_{\varrho}^l$ from $A^l$, when overlaid with the original image $x_i$, can be denoted as $\bar{A}_{\varrho}^l$ :

\begin{equation}
\bar{A}_{\varrho}^l=A_{\varrho}^l \otimes x_i 
\end{equation}
where, $\otimes$ represents element-wise multiplication. $\bar{A}_{\varrho}^l$ is then fed into a DCSCNs model $\Delta$, and the softmax function produces category scores $S_q^l$ :

\begin{equation}
 S_q^l=\operatorname{Softmax}\left(\Delta\left(\bar{A}_{\varrho}^l\right)\right) 
\end{equation}

The DCSCNs model $\Delta$ predicts the category result $y^q$ for $x_i$, where $q=[1, \cdots, m]$, and the category score for the $\varrho^{th}$ channel feature map is denoted as $S_{\varrho, q}^l$. The definition of the channel feature independence coefficient is as follows:
\begin{equation}
F C_{\varrho}^l=\frac{\left\|A^l\right\|_*-\left\|\Xi_{\varrho}^l \odot A^l\right\|_*}{\left\|A^l\right\|_*} 
\end{equation}
where, $FC_{\varrho}^l$ represents the independence score of each channel feature map, $ A^l \in \mathbb{R}^{H \times W \times C_l}$, and $A_{\varrho}^l \in \mathbb{R}^{H \times W}$ represents the $\varrho$, $\varrho \in\left[1, C_l\right]$ feature channel. $\|\cdot\|_*$ represents the nuclear norm, $\odot$ represents the Hadamard product, and $\Xi_{\varrho}^l$ represents the convolutional kernel mask matrix, where the $\varrho^{th}$ row is zero and the other rows are 1. By multiplying all the feature maps in $A^l$ by $S_{\varrho, q}^l$ and $F C_{\varrho}^l$, and then summing them, the CAM for sample $x_i$ under the $l^{th}$ layer DCSCNs model can be obtained:

\begin{equation}
\mathcal{L}^q=\operatorname{ReLU}\left(\sum_{\varrho} F C_{\varrho}^l S_{\varrho, q}^l A_{\varrho}^l\right) 
\end{equation}

Based on $\mathcal{L}^q$, the interpretable trustworthiness evaluation index can be defined to measure the deviation between the predicted target and the true target, determining whether the interpretability result is consistent with the true result. The interpretable trustworthiness evaluation index for the $i^{th}$ sample is defined as the intersection over union (IoU) between the highlighted region $d_i$ of the interpretability result and the true annotated region $\bar{d}_i$:
\begin{equation}
 {IoU}_i=\frac{d_i \cap \bar{d}_i}{d_i \cup \bar{d}_i} 
\end{equation}
where, $d_i \cap \bar{d}_i$ represents the area of intersection between $d_i$ and $\bar{d}_i$, and $d_i \cup \bar{d}_i$ represents the area of union between $d_i$ and $\bar{d}_i$. The interpretable trustworthiness evaluation index ${IoU}^l$ for the $l^{th}$ layer network can be obtained from ${IoU}_i$ :
\begin{equation}
{IoU}^l=\frac{\sum_{i=1}^N {IoU}_i}{N} 
\end{equation}
where, $N$ represents the number of samples.

\subsection{Adaptive pruning mechanism for convolutional kernels based on RL}

During the construction process of DCSCNs, the maximum number of convolutional kernels in each layer network is manually set. However, the correlations between these kernels may introduce redundant features \cite{9477103}. These features not only increase the complexity of the network but also reduce the computational efficiency. Therefore, the independent scores of the channel feature maps are used to evaluate the independence of feature map corresponding to each convolutional kernel. When a feature map has low independence with other channel feature maps, it means that its information has been largely encoded in other feature maps. Based on the independent scores of each channel feature map, the convolutional kernels can be sorted according to their scores. To enable the deep learning model to learn sparse representations of features, the DDPG (deep deterministic policy gradient) RL algorithm, as shown in Figure 4, is adopted for the adaptive pruning of the number of convolutional kernels \cite{mnih2015human}. A joint reward function based on recognition accuracy, interpretability reliability evaluation index, and parameter count is defined to balance the accuracy and interpretability reliability of the deep learning model. Through RL, the network width is adaptively adjusted to select feature maps with low correlations, thus improving the model's inference speed and reducing the risk of over-fitting. The settings of actions, states, rewards, and training strategies in the deep RL model are as follows:

\begin{figure}[!t]
\centering
\includegraphics[width=3in]{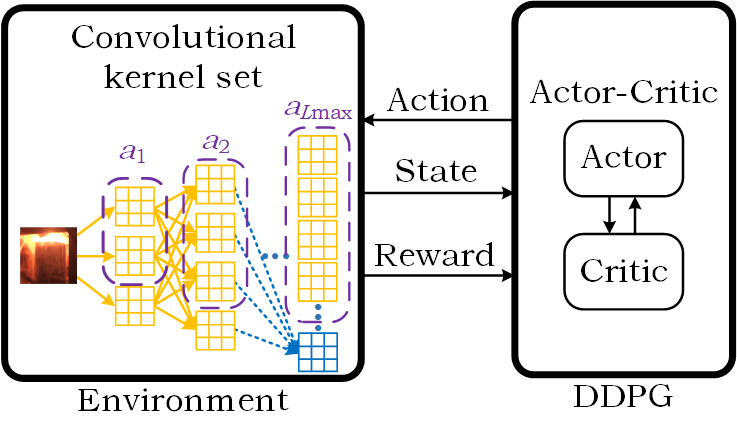}
\caption{Structure of convolutional kernel adaptive pruning based on feature map independence.}
\label{fig_4}
\end{figure}

Action Set: A continuous action set $A=\{a\}$ represents the pruning ratio of each layer convolutional kernels, where $a \in(0,1)$.

State Set: The state set of the $l^{th}$ layer convolutional layer is denoted as $S=\left\{l, C_l, C_{l-1}, h,  w, a_{l-1}\right\}$, where $l$ represents the current convolutional layer, $C_l$ represents the number of convolutional kernels in the current layer, $C_{l-1}$ represents the number of convolutional kernels in the previous layer, $h$ represents the length of the input feature map, $w$ represents the width of the input feature map, and $a_{l-1}$ represents the action of the previous layer.

Reward Function: To balance feature independence, model accuracy, and interpretable trustworthiness, the joint reward function is defined as follows:
\begin{equation}
 R=ACC_{val}+I o U_{val}-\beta \times P A 
\end{equation}
where, $ACC_{val}$ is the accuracy of the pruned model on the validation set, $IoU_{val}$ is the interpretable trustworthiness evaluation index of the final layer feature map in the pruned model on the validation set, $PA$ is the parameter count after pruning, and $\beta$ is the weight coefficient.

Training Strategy: The DDPG RL algorithm is used, where the actor network performs action selection, including the policy network $\mu$ and the target policy network $\mu^{\prime}$. The critic network evaluates values, including the value network $Q$ and the target value network $Q^{\prime}$. The objective function of the critic value network is:
\begin{equation}
Loss=\frac{1}{N} \sum_t^N\left(y_t-Q\left(s_t, a_t \mid \omega\right)\right)^2
\end{equation}
where, $N$ is the number of samples, $Q\left(s_t, a_t \mid \omega\right)$ is the value of evaluating state $s_t$ and action $a_t$ in the critic value network with parameters $\omega$, $y_t$ is the target value:
\begin{equation}
y_t=R_t+\gamma Q^{\prime}\left(s_{t+1}, \mu^{\prime}\left(s_{t+1} \mid \omega\right) \mid \omega^{\prime}\right)
\end{equation}
where, $\mu^{\prime}\left(s_{t+1} \mid \omega\right)$ is the action prediction for state $s_{t+1}$ at time $t+1$ using the actor target policy network, $\omega^{\prime}$ is the parameter of $Q^{\prime}$, and $\gamma$ is the discount factor.

The actor policy network is updated using policy gradients:
\begin{equation}
\begin{gathered}
\quad \nabla_{\theta^\mu} J \approx \\
\left.\left.\frac{1}{N} \sum_t \nabla_{\theta^{\mu}} \mu(s \mid \theta^{\mu})\right|_{s=s_t} \nabla_a Q(s, a \mid \omega)\right|_{s=s, a=\mu\left(s_t\right)}
\end{gathered}
\end{equation}
where $\theta^{\mu}$ is the parameter of the policy network $\mu$.

Update the target policy network $\mu^{\prime}$ and the target value network $Q^{\prime}:$
\begin{equation}
\theta^{\mu ^{\prime}}  \leftarrow \tau \theta^{\mu}+(1-\tau) \theta^{\mu ^{\prime}}
\end{equation}

\begin{equation}
\omega^{\prime}  \leftarrow \tau \omega+(1-\tau) \omega^{\prime}
\end{equation}
where, $\tau$ is the learning rate, $\tau \in(0,1)$.

\section{Experimental analysis}

\subsection{Data description and processing}

To verify the effectiveness of the proposed method, production videos were selected from an fused magnesium furnace factory in Liaoning Province. The videos were frame-separated to obtain images with a resolution of $1080\times 1920$. After image data augmentation, a total of 12,000 image samples were obtained. Figures 5 to 8 show the partial results of data augmentation for normal, underburn, overheated, and abnormal exhaust working conditions, respectively. The working conditions of the fused magnesium furnace images were annotated by experts. Randomly, 60\% of the data (7,200 images in total) were selected as the training set, with each of the four condition samples accounting for a quarter of the training set. Additionally, 20\% of the data (2,400 images in total) were chosen as the validation set, and the remaining 20\% (2,400 samples) were used as the test set. The experimental setup included an Intel i9-10900K processor, 16GB of memory, and an RTX 3060 graphics card.

\begin{figure*}[!t]
\centering
\includegraphics[width=6.5in]{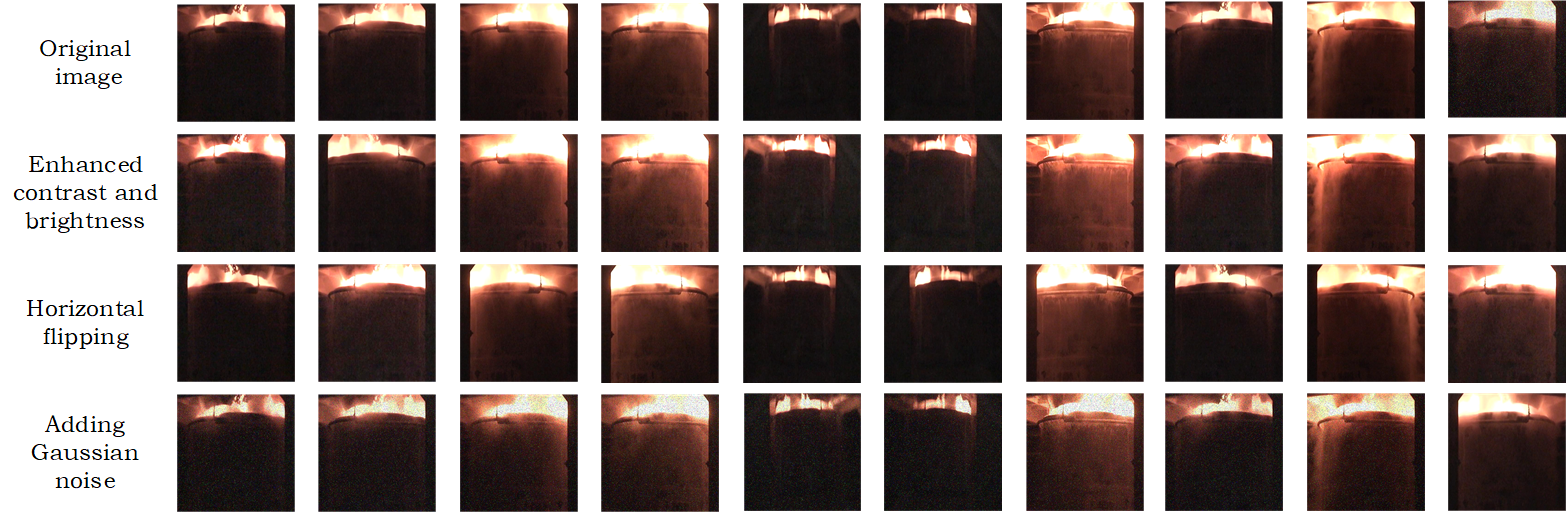}
\caption{ Results after image data enhancement for normal working conditions.}
\label{fig_5}
\end{figure*}

\begin{figure*}[!t]
\centering
\includegraphics[width=6.5in]{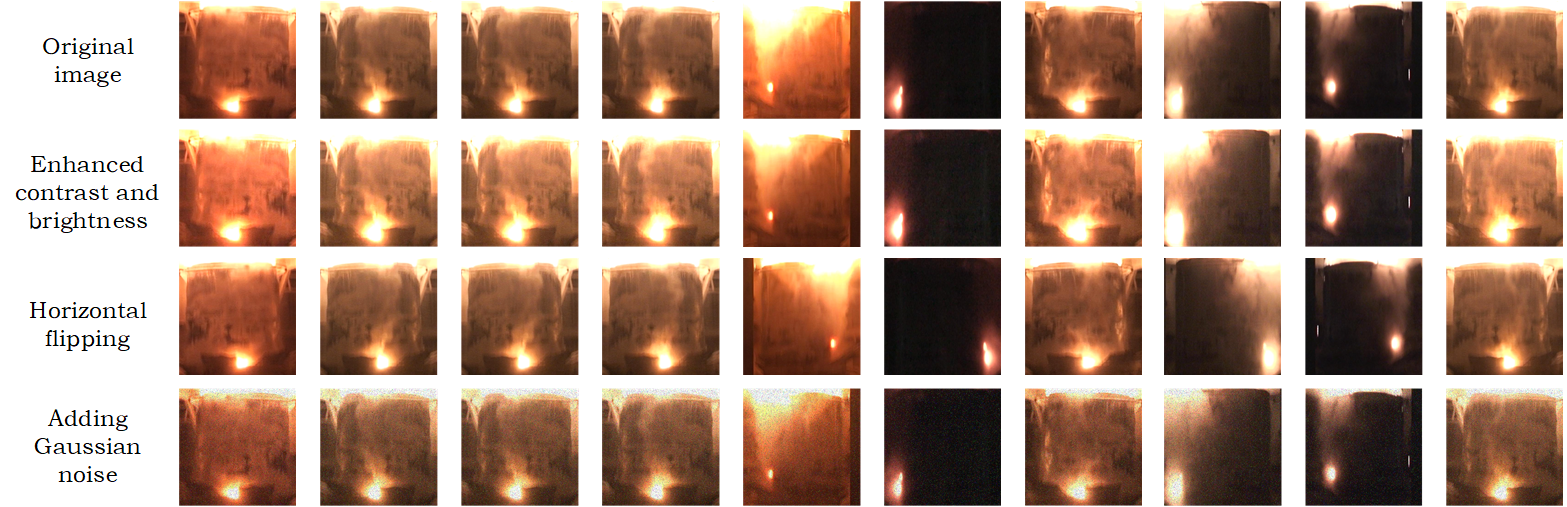}
\caption{Results after image data enhancement for underburn working conditions.}
\label{fig_6}
\end{figure*}

\begin{figure*}[!t]
\centering
\includegraphics[width=6.5in]{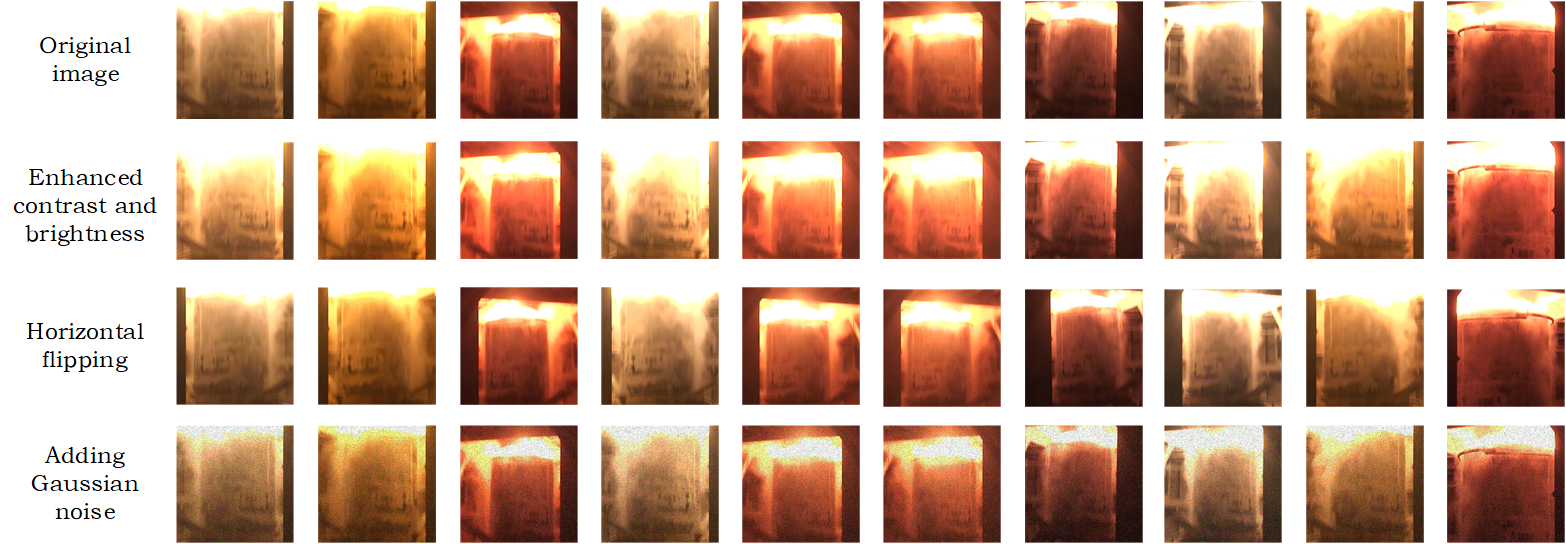}
\caption{ Results after image data enhancement for overheating working conditions.}
\label{fig_7}
\end{figure*}

\begin{figure*}[!t]
\centering
\includegraphics[width=6.5in]{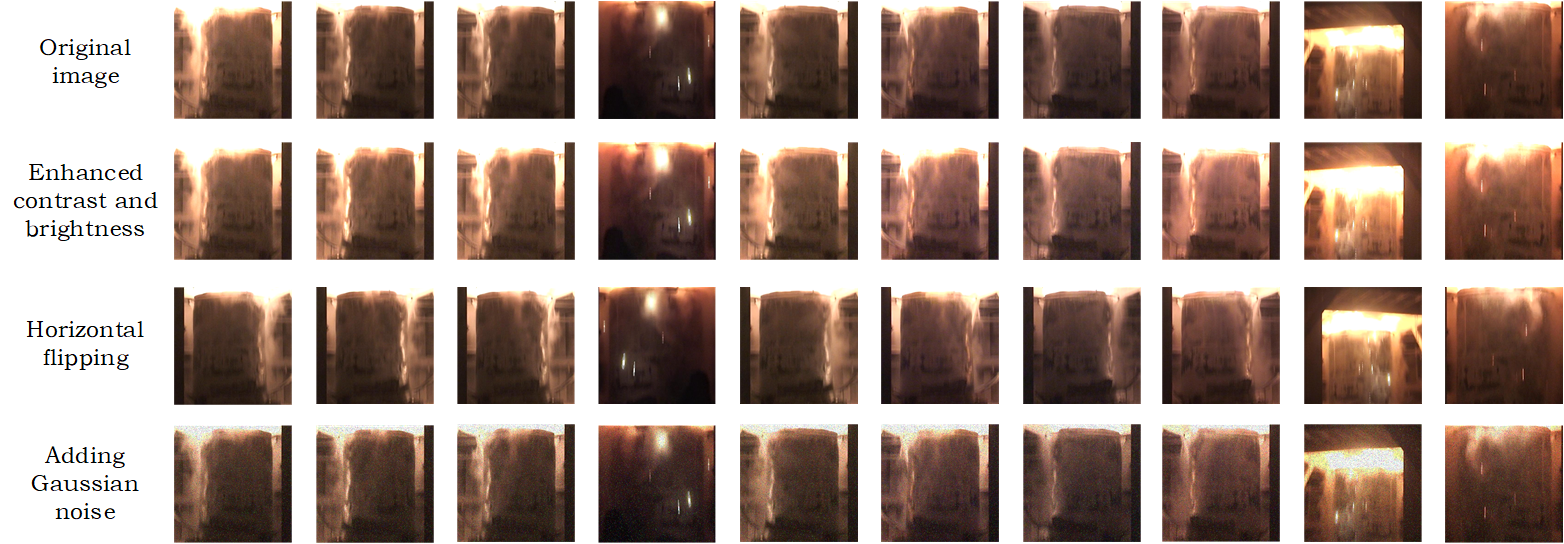}
\caption{Results after image data enhancement for abnormal exhaust working conditions.}
\label{fig_8}
\end{figure*}

\subsection{Performance evaluation metrics}

The accuracy $(R_a)$ is defined as the ratio of the number of correctly classified samples to the total number of samples, and it is calculated as follows:

\begin{equation}
R_a=\frac{T P+T N}{N}
\end{equation}
where, $T P$ represents the number of samples that are truly positive and are correctly predicted as positive, and $T N$ represents the number of samples that are truly negative and are correctly predicted as negative.

The parameter count $(PA)$ measures the memory size occupied by the model, which is the sum of the parameter count of the convolutional layers $(CA)$ and the fully connected layers $(FA)$, and it is calculated as follows:
\begin{equation}
\begin{aligned}
P A & =\sum_{i=1}^{L_{\max }}\left(C A_l+F A_l\right) \\
& =\sum_{i=1}^{L_{\max }}\left(k \times k \times C_{l-1}+1\right) \times C_l+\left(m_{i n}+1\right) \times m
\end{aligned}
\end{equation}
where, $C_{l-1}$ is the number of convolutional kernels in the $l-1^{th}$ layer, $C_l$ is the number of convolutional kernels in the $l^{th}$ layer, $m_{\text {in}}$ is the number of input neurons in the fully connected layer, and $m$ is the dimension of the network output.

\subsection{Experiment results and analysis}
\subsubsection{Experiment results of the deep convolutional stochastic configuration networks}

The parameter settings for the DCSCNs are as follows: the preferred error limit is $\bar{e}=0.01$, the maximum number of candidate Gaussian differential convolutional kernels is $T_{\max }=100$, the standard deviation $\xi$ ranges from 0.5 to 5 , the scale factor $r$ ranges from 0.8 to 1.5 , and the sizes of the Gaussian differential convolutional kernels is $k=\{3,5,7\}$. The maximum number of convolutional layers in the network is $L_{\max }=10$, the maximum number of convolutional kernels per layer is $C_{L_{\text {max }}}=50$, a sigmoid activation function is used, the pooling layer adopts the maximum pooling method with a kernel size of $k_p=2$, and a non-negative contraction sequence is defined as $u_C=\frac{1}{C}$, where ${C}$ ranges from 1 to $C_{\max }$. A pooling layer is inserted after every two convolutional layers.

Figure 9 shows the recognition accuracy curves of the training and testing samples of the fused magnesium furnace under different sizes of convolutional kernels when the network has only one convolutional layer $(l=1)$. It can be observed that during the training and testing processes, as convolutional kernels with different sizes and meaningful configurations are generated one by one, the DCSCNs model exhibits rapid convergence. After 50 convolutional operations, the model's accuracy reaches a plateau. When the number of convolutional kernels is too small, the network fails to extract sufficient information. On the other hand, as the number of convolutional kernels continues to increase, the improvement in network performance becomes limited while the computational complexity increases. Therefore, this study sets the maximum number of convolutional kernels per layer to 50. Since the performance of a single convolutional layer is limited, multiple convolutional layers are necessary to extract abstract features and achieve the desired classification performance. Additionally, while different sizes of convolutional kernels can achieve similar model performance, they also bring additional computational complexity. Two 3$\times$3 convolutional kernels have the same receptive field as one 5$\times$5 convolutional kernel, but the two 3$\times$3 kernels can perform two nonlinear transformations, providing better nonlinear transformation capability with lower computational complexity. Therefore, a kernel size of 3 can better balance model performance and complexity.

\begin{figure}[!t]
\centering
\includegraphics[width=3in]{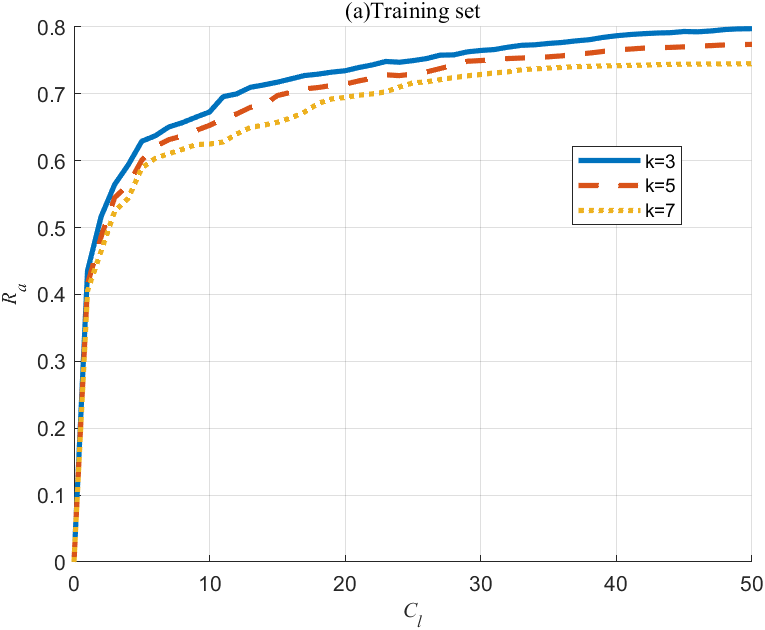}
\label{fig_9a}
\end{figure}

\begin{figure}[!t]
\centering
\includegraphics[width=3in]{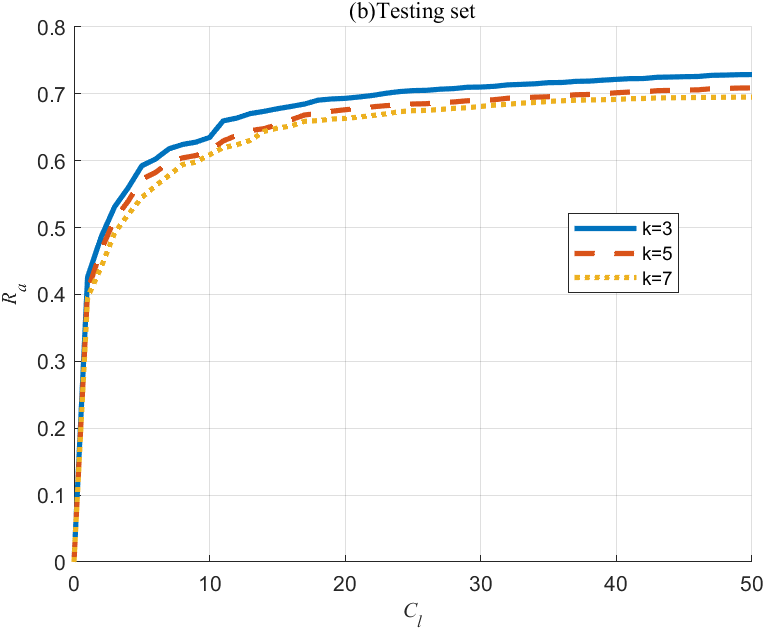}
\caption{Single-layer DCSCNs recognition accuracy curves with different convolutional kernel sizes.}
\label{fig_9b}
\end{figure}

By fixing the kernel size to $k=3$, Figure 10 shows the recognition accuracy curves of the training and testing sets for an 8-layer DCSCNs model. It can be observed that the multi-layer network exhibits better convergence compared to a single-layer network, with an accuracy of 94.53\% on the training set and 92.78\% on the testing set. The multi-layer DCSCNs achieves an improvement of 14.78\% and 19.89\% in accuracy compared to the single-layer DCSCNs on the training and testing sets, respectively, while the training time increases by 73.21\%. 

\begin{figure}[!t]
\centering
\includegraphics[width=3in]{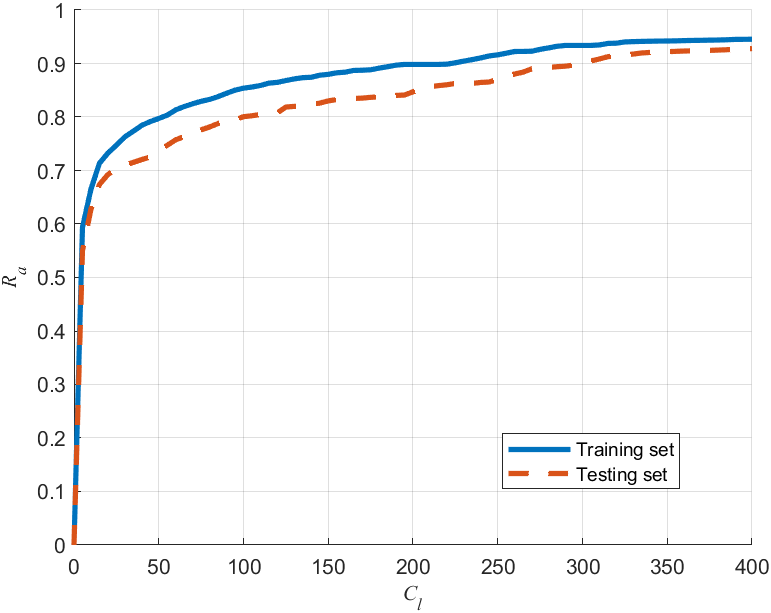}
\caption{Recognition accuracy curve of multi-layer DCSCNs network.}
\label{fig_10}
\end{figure}

\subsubsection{Results of the interpretability evaluation experiment}

Figure 11 depicts the class activation mapping for the four working conditions in the deep convolutinal stochastic configuration network, with varying numbers of convolutional layers. The highlighted regions of the figure denotes the activation region, which represents the area of focus for this convolutional layer. When $l=1$, the highlighted regions are small and dispersed, indicating that the network is unable to effectively focus on the target region. This suggests that the network is unable to extract sufficient feature information at this number of layers. When $l=4$, the highlighted region becomes larger and is in closer proximity to the target region, indicating an improvement in feature extraction. When $l$ is further increased to 8, the highlighted region encompasses the entire target region, indicating that as the number of convolutional layers increases and the amount of computation increases, the network is able to extract more critical and complete feature information. Therefore, in this paper, $l$ is set to 8.

The class activation mapping maps for the four working conditions under $l=8$ are presented in Fig. 12(a)-12(d). It can be observed that the highlighted regions with the various working conditions exhibit complete and continuous characteristics, allowing for the clear recognition of the target region. This suggests that when $l$ is 8, the DCSCNs are capable of adapting to the image data of different working conditions, effectively extracting the key features of each working condition, and accurately locating the target region of interest. Specifically, under normal working conditions, the network focuses on the position of the flame at the furnace opening. Under under-burning working conditions, the network is able to focus on the position of the furnace wall burning red and the furnace opening. Under overheating working conditions, the network accurately locates the position of the flame at the furnace opening. Under abnormal exhaust working conditions, the network is able to recognise the region where the solution overflows from the furnace opening. In conclusion, the proposed model provides reliable technical support for practical species for fused magnesium furnace working condition recognition.

\begin{figure}[!t]
\centering
\includegraphics[width=3.3in]{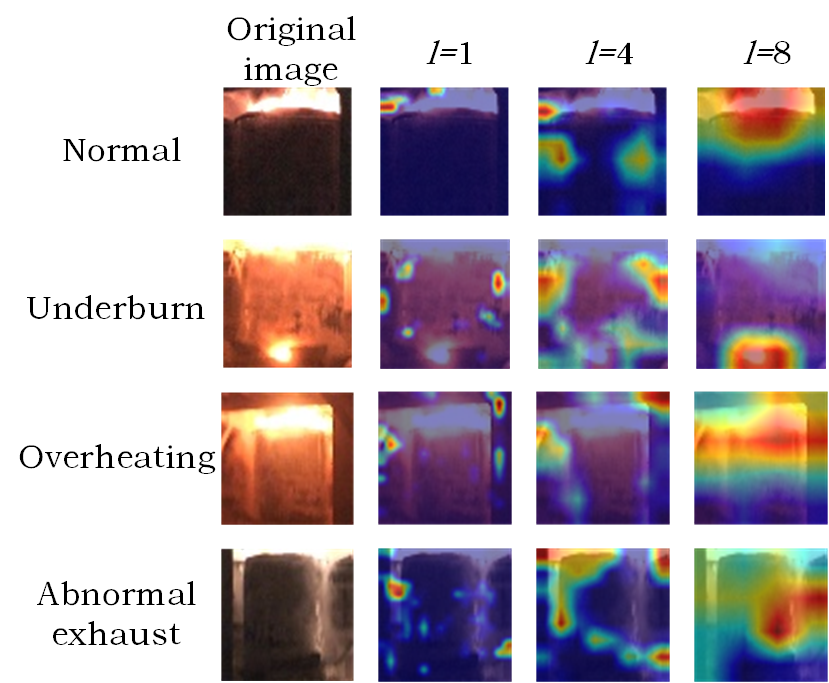}
\caption{Class activation mapping for different working conditions with different number of convolutional layers}
\label{fig_11}
\end{figure}

\begin{figure*}[!t]
\centering
\includegraphics[width=6in]{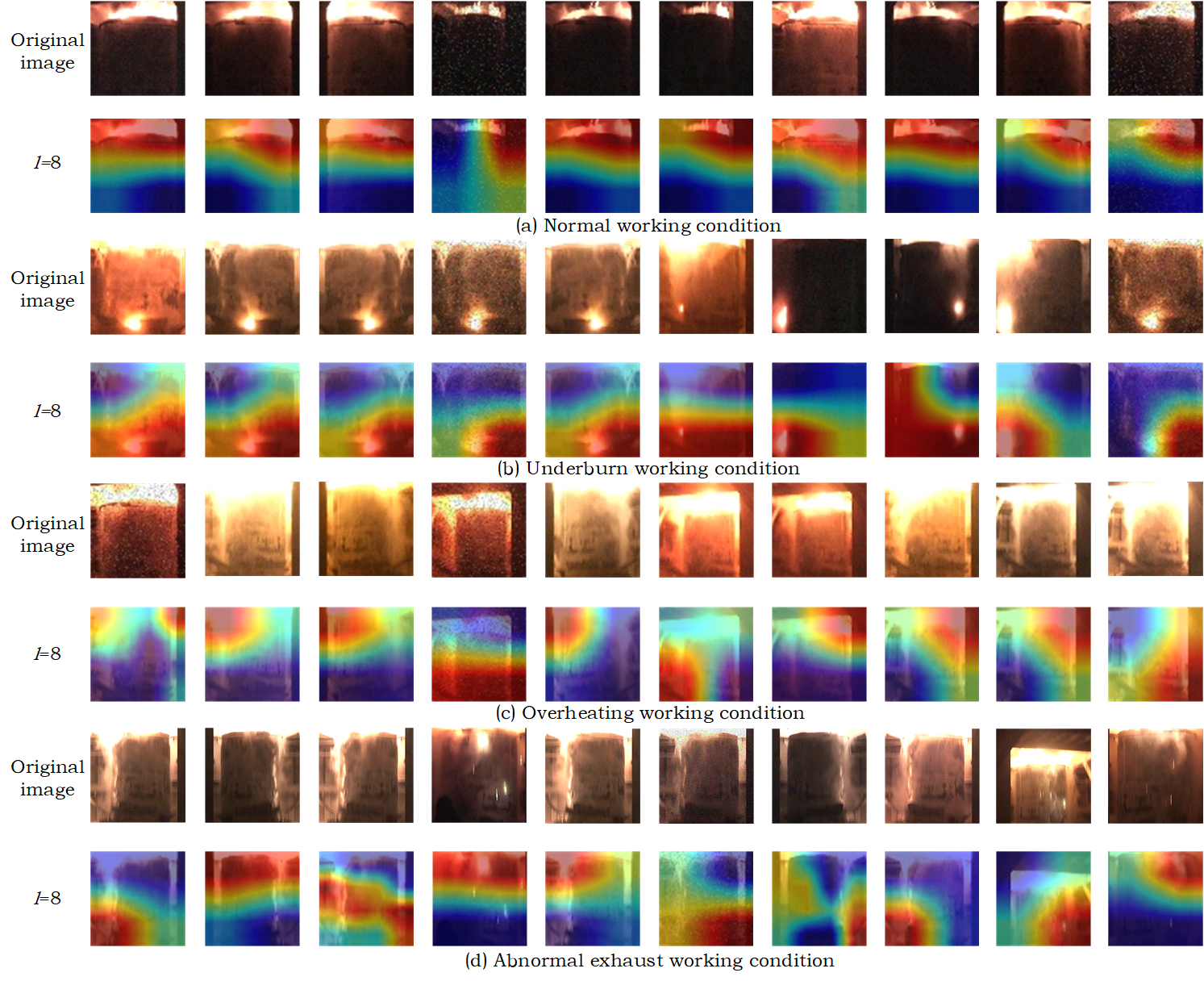}
\caption{Class activation mapping for different working conditions.}
\label{fig_12}
\end{figure*}

\subsubsection{Convolutional kernel adaptive pruning experimental results}

The parameter settings for the deep RL model are as follows: both the actor and critic networks are two-layer fully connected networks with 300 neurons, the discount factor $\gamma$ is set to 0.9, the experience replay pool size is 2000, the learning rate $\gamma$ is 0.005, and the training consists of 400 epochs.

Figure 13 shows the average reward curve after the Gaussian differential convolutional kernel set is fed into the RL module in the 8-layer DCSCNs. From this figure, it can be observed that after 300 training rounds, the average reward value stabilizes around 1.55, indicating the convergence of the RL module and the selection of a convolutional kernel set that balances model accuracy, interpretability, and parameter size.

\begin{figure}[!t]
\centering
\includegraphics[width=3in]{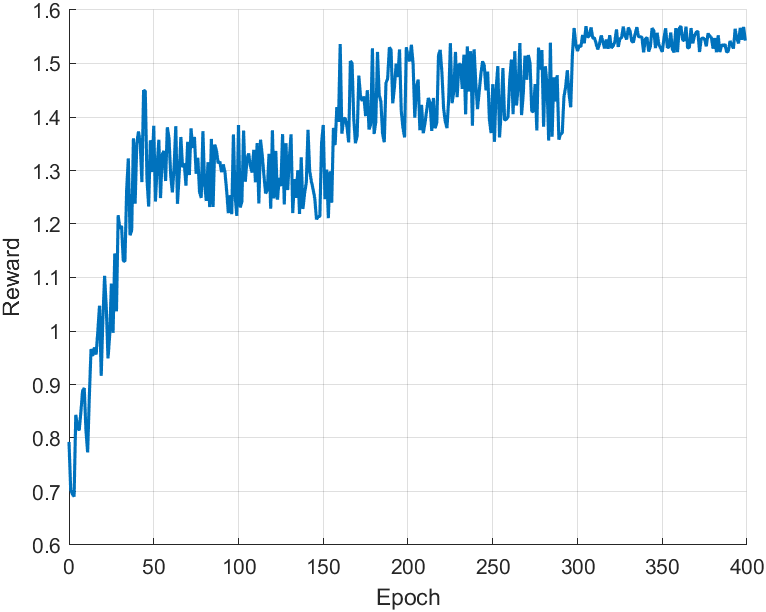}
\caption{The average reward curve of the RL training process.}
\label{fig_13}
\end{figure}

To further demonstrate the advantages of the proposed algorithm compared to the feature selection algorithm without RL, a comparative analysis of the interpretability of the models with and without the RL module is presented in Figure 14. From the figure, it can be observed that the proposed algorithm achieves more accurate and clear interpretability compared to the algorithm without RL. Specifically, the algorithm without RL exhibits unnecessary highlights in other parts of the image, despite having high attention on the flame of the furnace wall. These redundant highlights negatively impact the final decision, reducing the reliability and accuracy of the algorithm. In contrast, the proposed algorithm can accurately highlight the anomalous regions with higher precision and concentration, while reducing noise. This indicates that the features selected through RL exhibit higher independence and effectively eliminate redundant features, resulting in a more accurate and reliable algorithm. This advantage not only enhances the algorithm performance but also makes the internal workings of the algorithm more transparent and easier to understand, meeting the requirement for interpretability in industrial production.

\begin{figure}[!t]
\centering
\includegraphics[width=3.5in]{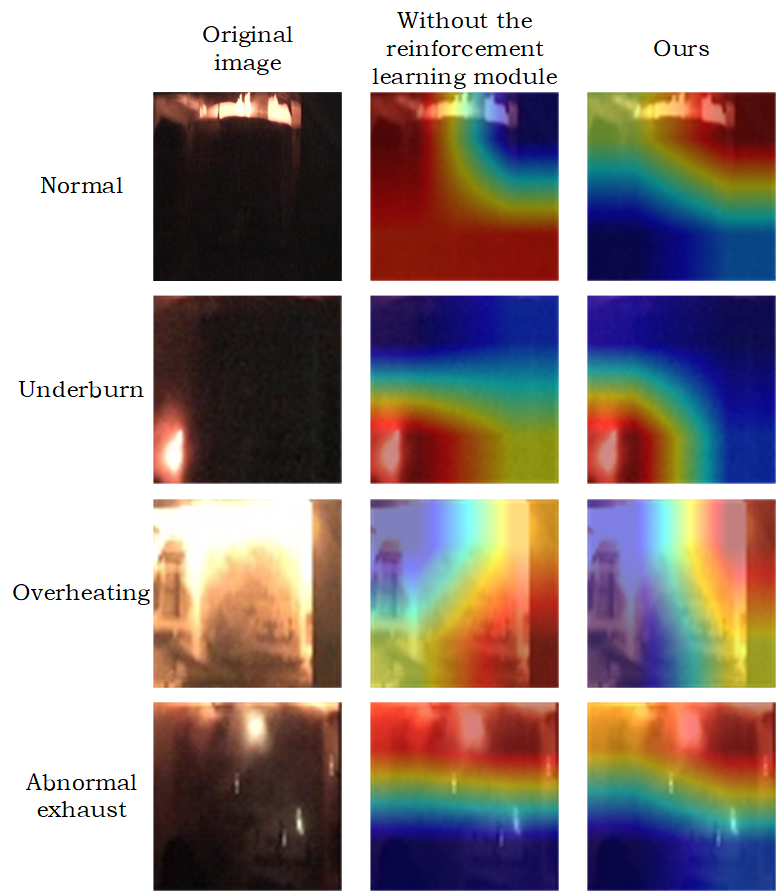}
\caption{Comparison of the algorithm proposed in this paper with the class activation mapping without RL algorithm.}
\label{fig_14}
\end{figure}

\subsection{Ablation Experiments}

To validate the effectiveness of each module proposed in this study, a comparative analysis is conducted against three variations: without Gaussian convolutional kernels, without the adaptive kernel pruning mechanism, and without both Gaussian convolutional kernels and the adaptive kernel pruning mechanism. An 8-layer DCSCNs is employed for these experiments. The results are presented in Table 1.

\begin{table*}[htbp]
\caption{Ablation experiment results}
\begin{center}
\begin{tabular}{cccccc}
\toprule 
Model & Training $R_a$/\% & Testing $R_a$/\% & $PA$/MB &  Training time$/s$	& Inference time$/s$\\
\midrule 
Proposed method &93.96&92.57&\textbf{48.85}&19338.943&\textbf{0.014}\\   
Without Gaussian convolutional kernels &93.04&91.86&\textbf{48.85}&\textbf{17157.832}&\textbf{0.014}\\
Without adaptive kernel pruning mechanisms &\textbf{94.53}&\textbf{92.78}&80.25&31475.854&0.019\\
Without Gaussian convolutional kernels\\ and adaptive kernel pruning mechanisms&93.38&92.01&80.25&23656.331&0.016\\
\toprule 
\end{tabular}
\label{tab2}
\end{center}
\end{table*}

As can be inferred from Table 1, the proposed algorithm shows improvements in both training and testing accuracy rates compared to the methods without Gaussian convolutional kernels and those also without the adaptive kernel pruning mechanism module. Specifically, the training and testing accuracy rates are enhanced by 0.92\%, 0.58\%, 0.71\%, and 0.56\%, respectively. Furthermore, compared to the method lacking the adaptive kernel pruning mechanism, our approach reduced the parameter size by 31.4MB, and the training and inference time are decreased by 12136.911s and 0.005s, respectively. Therefore, the utilization of meaningful Gaussian differential convolutional kernels enables the model to better adapt to the complex scenarios encountered in fused magnesium furnace working conditions, thereby ensuring continual improvements in accuracy. CAM is employed for visualizing feature importance, and the interpretable trustworthiness index is defined, thereby making the working condition recognition more accurate and trustworthy. Moreover, the adaptive pruning mechanism proposed in this algorithm manages to reduce the loss in accuracy within an acceptable range, effectively reducing the parameter size and training time, and subsequently enhancing inference speed.

\subsection{Performance Comparison Experiment}

To validate the effectiveness of the proposed algorithm, it is compared with the performance of the SCNs \cite{8013920}, 2DSCNs \cite{LI2024112249}, DeepSCNs \cite{8489695}, CNNs \cite{9451544}, Bayesian Network, and CNNs+LSTM working condition recognition models. For SCNs and 2DSCNs, the number of hidden layers is set to 1 and the number of nodes in the hidden layer is set to 2000. For DeepSCNs, the number of hidden layers is 4, and each hidden layer has 500 nodes. The activation function used is sigmoid. The range of the hidden layer node parameters $\lambda $ is $\left\{1, 3, 5, 7, 9, 10, 25, 50, 100 \right\}$, $\varepsilon$ is set to 0.01, and the contraction sequence $r$ is $\left\{0.9, 0.99, 0.999, 0.9999, 0.99999, 0.999999\right\}$. The CNNs model consists of 8 convolutional layers, 8 sigmoid activation layers, 4 pooling layers, and 1 fully connected layer, and it is trained for 100 epochs.

Figure 15 shows the training recognition accuracy curves of different models. From the figure, it can be observed that the training recognition accuracy curves of the four models converge as the number of convolutional kernels/hidden layer nodes increases. SCNs, 2DSCNs, and DeepSCNs converge slowly around 2000 hidden layer nodes with recognition accuracies of approximately 0.77, 0.79, and 0.84, respectively. In contrast, the proposed method exhibits more dramatic changes in recognition accuracy, reaching 0.94 after 400 convolutional operations.

\begin{figure}[!t]
\centering
\includegraphics[width=3in]{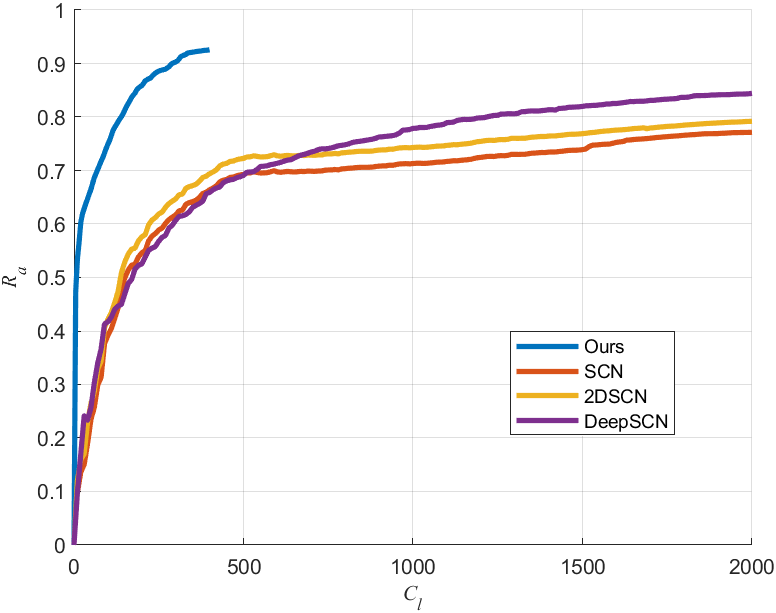}
\caption{Recognition accuracy curves of training samples for different network models.}
\label{fig_15}
\end{figure}

Table 2 presents a comparison of the testing recognition accuracy, parameter size, training time, and inference time between the proposed algorithm and the other randomized methods. The proposed algorithm achieves the highest accuracy while maintaining the smallest number of parameters compared to SCNs, 2DSCNs, and DeepSCNs. In terms of training time, the multi-layer models of DCSCNs and DeepSCNs require more training time compared to the single-layer models of SCNs and 2DSCNs, while the inference time is at a similar order of magnitude. These findings suggest that the hidden layer nodes in SCNs, 2DSCN, and DeepSCNs exhibit constrained capabilities in feature extraction when processing complex image data directly. The proposed algorithm effectively extracts features of different working conditions in the fused magnesium furnace, thereby ensuring the convergence of recognition errors. This markedly enhances the accuracy of recognition outcomes, diminishes complexity and mitigates the risk of model overfitting. With regard to parameter size, the advantage of parameter sharing in the convolutional operations of the proposed algorithm, in conjunction with multiple convolutions and downsampling operations in deep networks, results in a reduction of the size of the feature maps input to the fully connected layer by one to two orders of magnitude compared to SCNs, 2DSCNs, and DeepSCNs. Furthermore, the proposed method employs three-channel colour images as input and replaces the hidden layer nodes with convolutional operations, resulting in a longer training time in comparison to SCN, 2DSCN, and DeepSCNs, which utilize grayscale images as input.

\begin{table*}[htbp]
\caption{Performance comparison of different randomized methods}
\begin{center}
\begin{tabular}{ccccc}
\toprule 
Model & Testing $R_a$/\% & $PA$/MB & Training time$/s$	& Inference time$/s$\\
\midrule 
SCNs &76.14&500.038&\textbf{10278.834}&\textbf{0.011} \\
2DSCNs &77.99&1000.038&12352.771&0.013  \\
DeepSCNs &83.64&127.899&15411.081&0.013  \\
Proposed method &\textbf{92.57}&\textbf{12.854}&18218.021&0.015\\
\toprule 
\end{tabular}
\label{tab2}
\end{center}
\end{table*}

Moreover, Table 3 provides a comparative analysis of between the proposed algorithm and other fused magnesium furnace working condition recognition methods, including CNNs, Bayesian networks, Swin Transformer, and CNNs+LSTM. The proposed algorithm achieves accuracy improvements of 2.22\%, 2.65\%, 3.81\%, and 3.00\% over these methods, respectively. However, it exhibits an increase in parameter size by 12.136 MB, 12.769 MB, and 8.727 MB compared to CNNs, Swin Transformer, and CNNs+LSTM, respectively. The proposed method has the shortest training time, while its inference time remains comparable to other methods. This demonstrates that the proposed methods, which employ a randomized configuration of Gaussian differential convolutional kernel parameters with physical meanings based on a supervised learning mechanism, are capable of effectively extracting the features of different working conditions of a fused magnesium furnace. Furthermore, the definition of an interpretable trustworthiness index and the use of adaptive width adjustment through RL facilitate the attainment of optimal working conditions, thereby enhancing the recognition results. Since each feature map collection is connected to the output layer, the accuracy is enhanced, although the number of parameters is higher compared to CNN, Swin Transformer, and CNN+LSTM. Unlike these methods, which are trained using the gradient descent backpropagation method, the proposed approach utilizes an incremental convolutional kernel generation process driven by supervised learning. This approach effectively circumvents the potential issues associated with weight initialization, local minimums, and sensitivity to learning rate, significantly reducing the training time.

\begin{table*}[htbp]
\caption{Performance comparison of different fused magnesium furnace working condition recognition methods}
\begin{center}
\begin{tabular}{ccccc}
\toprule 
Model & Testing $R_a$/\% & $PA$/MB & Training time$/s$	& Inference time$/s$\\
\midrule 
CNNs&90.35&0.718&20714.322&0.015\\
Bayesian network&89.92&-&-&-\\
Swin transformer&88.76&\textbf{0.085}&31123.652&0.016\\
CNNs+LSTM&89.57&4.127&20159.642&0.015\\
Proposed method &\textbf{92.57}&12.854&\textbf{18218.021}&0.015\\
\toprule 
\end{tabular}
\label{tab2}
\end{center}
\end{table*}

\section{Conclusion}

To address the issues of poor generalization and weak interpretability in existing methods for working condition recognition in fused magnesium furnaces, this paper proposes an interpretable working condition recognition model. The innovations of this model are as follows:

1) Based on a supervised learning mechanism, an incremental stochastic configuration strategy is adopted to generate Gaussian differential convolutional kernels with physical meanings, which are used to construct a deep convolutional network. This ensures that the recognition errors converge gradually and the model exhibits transparency and interpretability in its network structure.

2) The class activation mapping method is employed to analyze the interpretability of the model, recognizing the furnace features that require attention. Interpretable trustworthiness evaluation metrics are defined, a joint reward function is constructed, and RL is used for the adaptive pruning of the convolutional kernels to obtain the optimal network width.

3) The proposed method achieves a test recognition accuracy of 92.57\% for working condition in fused magnesium furnaces, outperforming other recognition methods.

In the future, stochastic configuration machines (SCMs) \cite{wang2023stochastic} will be employed to further improve the modeling quality in terms of the models's light weight and interpretability.

\bibliographystyle{IEEEtran}
\bibliography{SCCN}

\vspace{-25pt} 

\begin{IEEEbiography}
[{\includegraphics[width=1in,height=1.25in,clip,keepaspectratio]{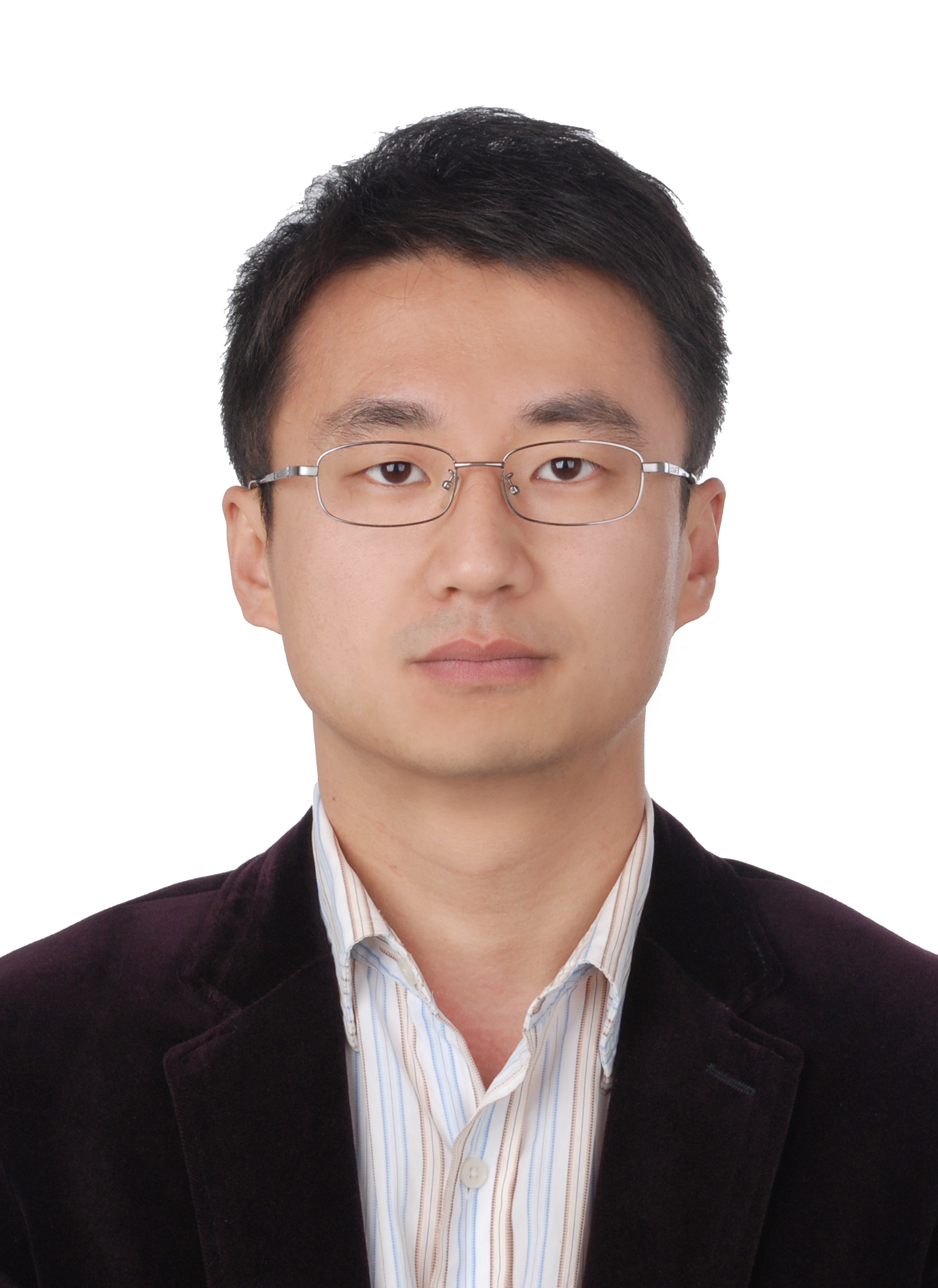}}] 
\noindent
\textbf{Weitao Li} received the Ph.D. degree is industrial automation from Northeastern University, Shenyang, China, in 2012. Since 2012, he has been an Associate Professor with the School of Electrical Engineering and Automation, Hefei University of Technology. His research interests include image processing, pattern recognition, and deep stochastic configuration networks for big data analytics. Dr. Li is an Associate Editor for Industrial Artificial Intelligence.
\end{IEEEbiography}

\vspace{-34pt} 

\begin{IEEEbiography}
[{\includegraphics[width=1in,height=1.25in,clip,keepaspectratio]{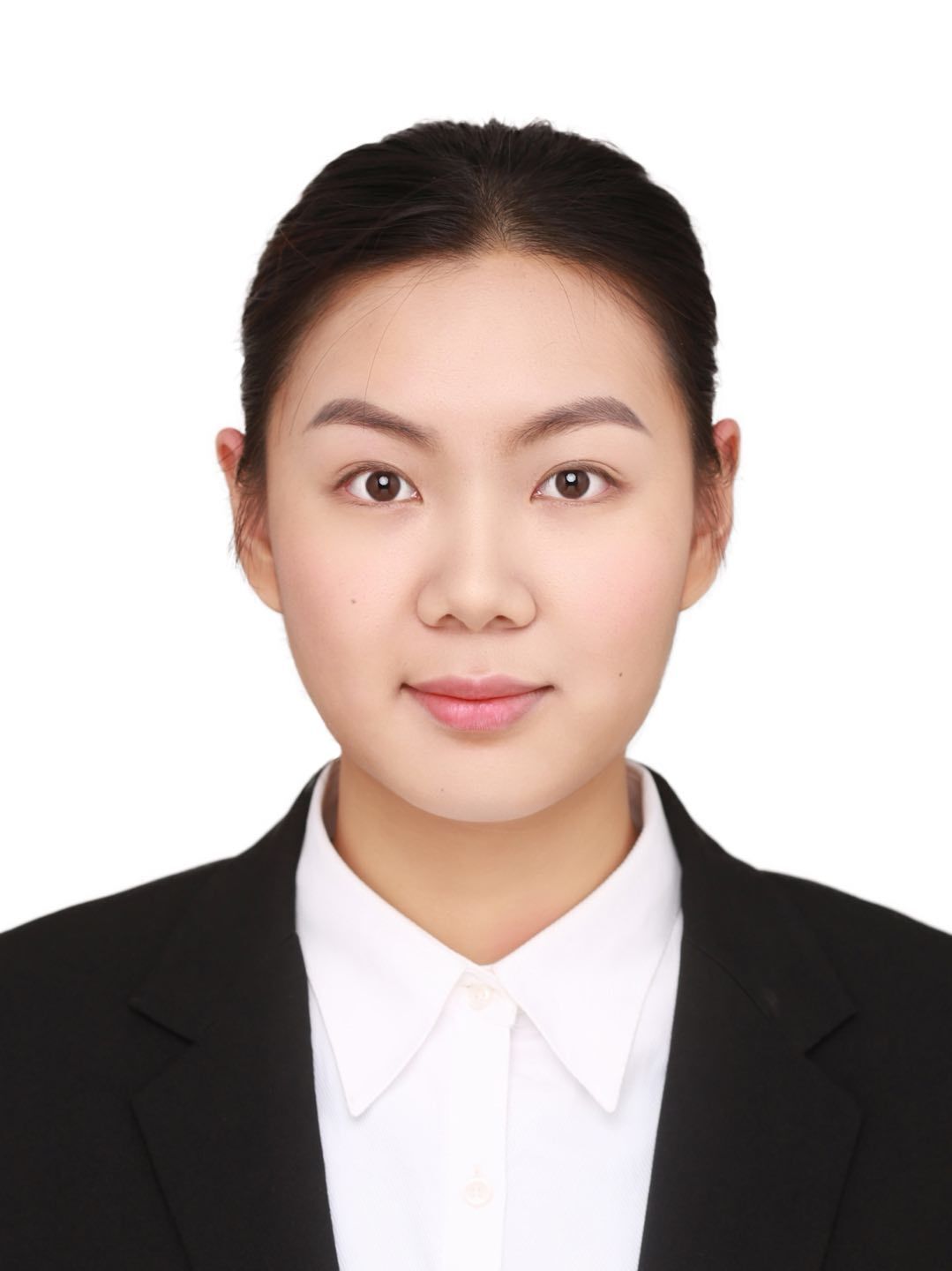}}] 
\noindent
\textbf{Xinru Zhang} received the B.S. degree in automation from Hunan University, Hunan, China, in 2018. She is currently pursuing a Ph.D. in Control Science and Engineering as a Master's-Ph.D. combined student at the School of Electrical Engineering and Automation, Hefei University of Technology. Her research interest covers deep learning, image processing, and industrial artificial intelligence.

\end{IEEEbiography}

\vspace{-35pt} 
\begin{IEEEbiography}
[{\includegraphics[width=1in,height=1.25in,clip,keepaspectratio]{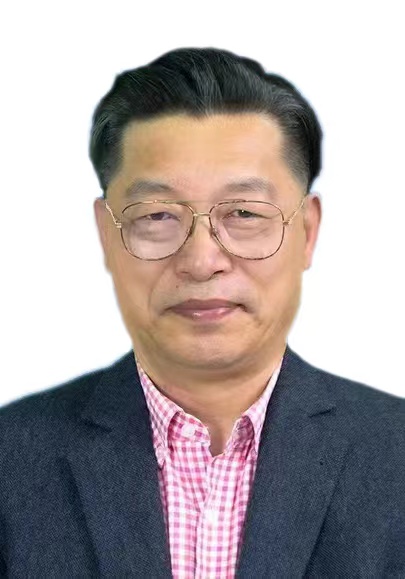}}] 
\noindent
\textbf{Dianhui Wang} (Senior Member, IEEE) received the Ph.D. degree in industrial automation from Northeastern University, Shenyang, China, in 1995. 
 From 1995 to 2001, he was a Postdoctoral Fellow with Nanyang Technological University, Singapore, and a Researcher with The Hong Kong Polytechnic University, Hong Kong. He was with La Trobe University, Australia, in 2001 and was a Reader and an Associate Professor with the Department of Computer Science and Information Technology until the end of 2020. Since 2017, he has been a Visiting Professor with the State Key Laboratory of Synthetical Automation of Process Industries, Northeastern University, China. Since July 2021, he has been with the AI Research Institute, China University of Mining and Technology, Xuzhou, China, working as the Dean of the AI Research Institute, a Distinguished Professor of CUMT, and the Director of the Research Center for Stochastic Configuration Machines. He is the primary funder of stochastic configuration networks and stochastic configuration machines, and the key advocate and practitioner of these randomized learner models in industrial data modeling, intelligent sensing system design, and other domain applications. 
 
 Dr. Wang is the Editor-in-Chief for Industrial Artificial Intelligence, and an Associate Editor for IEEE TRANSACTIONS ON CYBERNETICS, IEEE TRANSACTIONS ON FUZZY SYSTEMS, Information Sciences, Artificial Intelligence Review, and WIREs Data Ming and Knowledge Discovery.
\end{IEEEbiography}

\vspace{-30pt} 

\begin{IEEEbiography}
[{\includegraphics[width=1in,height=1.25in,clip,keepaspectratio]{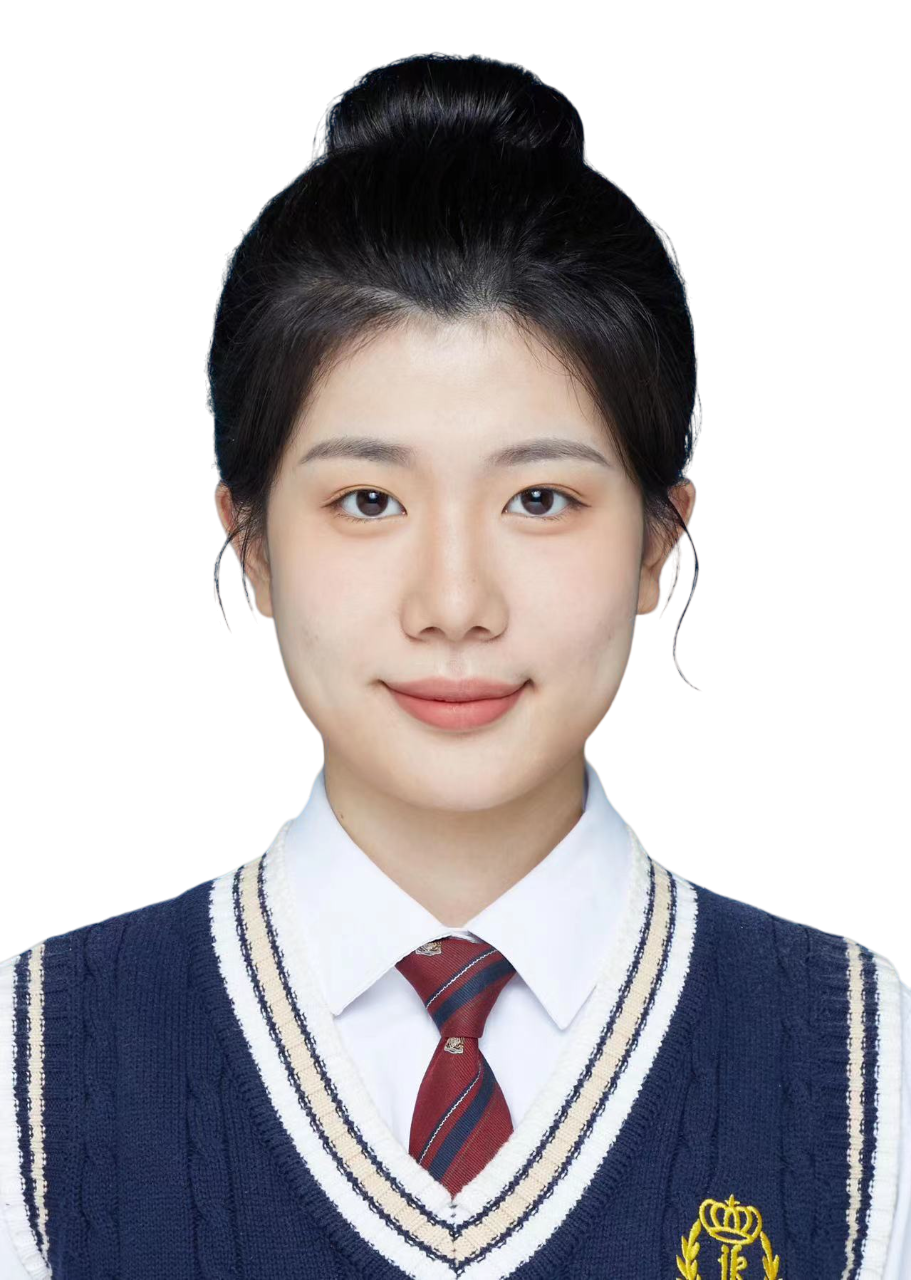}}] 
\noindent
\textbf{Qianqian Tong} received the B.S. degree in automation from Hefei University, Hefei, China, in 2017. She is currently pursuing the M.S.degree in Control Science and Engineering at the School of Electrical Engineering and Automation, Hefei University of Technology. Her research interest covers intelligent cognition, adaptive dynamic programming and optimal control.
\end{IEEEbiography}

\vspace{-19pt} 

\begin{IEEEbiography}
[{\includegraphics[width=1in,height=1.25in,clip,keepaspectratio]{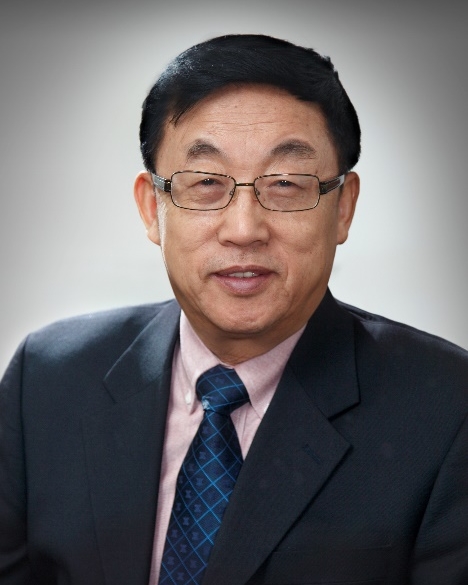}}] 
\noindent
\textbf{Tianyou Chai} (Life Fellow, IEEE) received the Ph.D. degree in control theory and engineering from Northeastern University, Shenyang, China, in 1985. He has been a Professor with Northeastern University since 1988. He is the Founder and Director of the Center of Automation, which became a National Engineering and Technology Research Center and a State Key Laboratory. He has published 150 international journal papers. His current research interests include modeling, control, optimization and integrated automation of complex industrial processes. Prof. Chai is a Member of the Chinese Academy of Engineering, an IFAC and IEEE Fellow, the Director of the Department of Information Science of the National Natural Science Foundation of China. For his contributions, he has won four prestigious awards of National Science and Technology Progress and National Technological Innovation, the 2007 Industry Award for Excellence in Transitional Control Research from the IEEE Multiple-Conference on Systems and Control.  His current research interests include modeling, control, optimization, and integrated automation of complex industrial processes.
\end{IEEEbiography}

\end{document}